\documentclass[lettersize,journal]{IEEEtran}
\usepackage{amsmath,amsfonts}
\usepackage{algorithmic}
\usepackage{algorithm}
\usepackage{array}
\usepackage{booktabs}
\usepackage{tabularx}
\usepackage{tcolorbox}
\tcbuselibrary{breakable,skins}
\usepackage[caption=false,font=normalsize,labelfont=sf,textfont=sf]{subfig}
\usepackage{textcomp}
\usepackage{stfloats}
\usepackage{url}
\usepackage{verbatim}
\usepackage{graphicx}
\usepackage{cite}
\usepackage[colorlinks=true,linkcolor=blue,citecolor=blue,urlcolor=blue]{hyperref}
\begin{document}

\title{EORestore-Agent: Fidelity-Guided Agentic Restoration of Remote Sensing Images with Composite Degradations}

\author{Heli~Qi,
Zeqi~Zhou,
Jingjun~Yi,
Kunyi~Liu,
Ziyang~Lihe,~\IEEEmembership{Student Member,~IEEE,}
Junjue~Wang,
Osamu~Yoshie,~\IEEEmembership{Member,~IEEE,}
and~Naoto~Yokoya,~\IEEEmembership{Member,~IEEE}%
\thanks{This work was supported by JST BOOST, Japan Grant Number JPMJBS2429, and by KAKENHI 26K21244. \textit{(Corresponding author: Osamu Yoshie (e-mail: yoshie@waseda.jp).)}}%
\thanks{Heli Qi, Kunyi Liu, and Osamu Yoshie are with the Graduate School of Information, Production and Systems, Waseda University, Kitakyushu, Fukuoka 808-0135, Japan.}%
\thanks{Zeqi Zhou, Junjue Wang and Naoto Yokoya are with the Graduate School of Frontier Sciences, The University of Tokyo, Kashiwa, Chiba 277-8561, Japan.}%
\thanks{Jingjun Yi is with the Informatics Institute, University of Amsterdam, Amsterdam 1098 XH, The Netherlands.}%
\thanks{Ziyang Lihe is with Wuhan University, Wuhan 430072, China.}%
\thanks{Heli Qi, Kunyi Liu, Junjue Wang, and Naoto Yokoya are also with the RIKEN Center for Advanced Intelligence Project (AIP), RIKEN, Tokyo \mbox{103-0027}, Japan.}}

\pagestyle{plain}


\maketitle
\thispagestyle{plain}

\begin{abstract}

Remote sensing images often carry composite degradations, in which haze, cloud, noise, blur, low light, and low resolution coexist. Restoring them requires deciding which tool to apply, in what order, and when to stop, yet no clean reference is available at inference time to verify these decisions. All-in-one models trained on single degradations converge to a narrow PSNR band as degradations accumulate.
To formulate real-world remote sensing restoration as a traceable trajectory, we present EORestore-Agent, which replaces this unmeasurable objective with reference-free, verifiable per-step decisions. A fine-tuned vision-language model reports all residual degradation types, whose tool pools are scored together, so the restoration order emerges from step-wise selection. A relative quality scorer, trained with full-reference supervision on synthetic degradation chains, predicts the changes in PSNR, SSIM, and LPIPS from the current image to each candidate. A step is accepted only when no predicted change is negative and the predicted PSNR gain is positive. Otherwise, the agent keeps the current image. On a synthetic Landsat-8 benchmark with six degradation types, EORestore-Agent improves PSNR by 2.3 to 3.2~dB over the strongest retrained all-in-one baseline on composites of two to six degradations, whereas zero-shot natural-image agents fall below the degraded input in PSNR in 17 of 18 settings. Replacing the learned scorer with no-reference quality differences costs 1.1 to 4.6~dB. The remaining harmful steps are small and cluster near the acceptance threshold. Sentinel-2 examples illustrate transfer to real atmospheric degradation without retraining.

\end{abstract}


\section{Introduction}
\label{sec:intro}

Remote sensing images acquired by satellite and airborne platforms are inevitably subject to multiple degradation processes that arise from diverse physical sources, including atmospheric scattering and absorption, variable illumination conditions, sensor noise characteristics, platform motion, and the intrinsic physics of the imaging chain~\cite{rasti2021image,zhou2026remote,kabir2020vicarious}. In practice, such degradation processes do not occur in isolation but act simultaneously and interact throughout the long-range Earth-surface-to-sensor optical path, giving rise to what is termed composite degradation, the coexistence of two or more heterogeneous degradation processes within a single observed image~\cite{cheng2026prompthsi,wang2024unified,feng2025progressive}. For example, a satellite scene may simultaneously exhibit haze from atmospheric scattering, cloud occlusion, and sensor noise, each governed by distinct physical mechanisms yet jointly corrupting the same acquisition.

To address diverse degradation types, all-in-one (AIO) image restoration has emerged as a prominent paradigm that trains a single unified model to handle multiple restoration tasks such as dehazing, denoising, deblurring, and low-light enhancement~\cite{potlapalli2023promptir,lihe2025ada4dir,dong2026phydae}. However, AIO models are fundamentally designed for multi-task capability: one model covering multiple degradation types, yet each training and test sample typically contains only a single degradation. As noted by Wang et al.~\cite{wang2025dual}, such one-to-many approaches can handle various degradation types individually but still restore only one degradation at a time. This assumption conflicts with real-world remote sensing scenarios where multiple degradations coexist in the same acquisition, demanding not only multi-task coverage but also the ability to perceive, disentangle, and sequentially or jointly address their combinations within a single image.

Composite degradation restoration introduces a set of decision variables beyond what a single restoration model can address: which degradation types are present, how they interact, in what order they should be addressed, whether to restore them jointly or sequentially, which specialized tool to invoke at each step, and when to terminate~\cite{chen2024restoreagent}. Among these decisions, how the restoration is sequenced is particularly consequential. Huang et al.~\cite{huang2025task} demonstrate that decomposing composite degradation into ordered subtasks and restoring them step by step yields substantial improvement over directly restoring the mixed degradation in a single pass. These findings suggest that composite restoration is not merely a harder restoration task but a multi-step decision problem, where the core challenge lies in which tool to apply at each step and when to stop.

Restoration agents offer a natural framework for such multi-step decision problems by decomposing the restoration process into a series of steps including degradation perception, strategy planning, and tool invocation. A growing body of work has validated this framework on natural images, employing large language or vision-language models to orchestrate specialized restoration tools in a multi-step pipeline~\cite{chen2024restoreagent,zhu2025intelligent,zhou2025q,zuo20254kagent,jiang2026multi,lin2025jarvisir}. These efforts confirm that the agentic workflow generalizes across different degradation combinations and tool configurations. However, the concrete components of these agents, from perception modules and toolboxes to quality criteria, are built around the degradation space of natural images. Directly transferring them to remote sensing introduces systematic domain gaps. Remote sensing involves degradation types absent from natural-image benchmarks, such as cloud occlusion~\cite{zheng2020single,zou2024diffcr}. Shared phenomena can carry different meanings: snow is a visibility-reducing degradation in natural images but valuable ground-cover information in remote sensing~\cite{liu2024fsc}. Even for common degradations, the underlying physics differs. Remote sensing dehazing must account for wavelength-dependent scattering across spectral bands~\cite{guo2020landsat}, whereas natural-image models assume camera-to-scene scattering alone~\cite{he2010single}. These gaps affect degradation perception, tool selection, and quality assessment alike, requiring a restoration agent designed for the remote sensing domain.

In designing such an agent, three coupled challenges inherent to sequential composite restoration must be addressed. First, the order in which degradations are restored significantly affects the final outcome~\cite{zhu2025intelligent}, and because restoration steps are not invertible, an unfavorable order committed at an early step cannot be undone later. Second, existing restoration agents typically assume a one-to-one mapping between detected degradation types and restoration tools~\cite{zhu2026opera}, so each step searches only the tools of a single predicted type. The truly best action can fall outside this truncated set, and Zhu et al.~\cite{zhu2026opera} observe that a majority of high-quality restoration trajectories include tools outside the scope of the detected degradation. Third, the verification signal that accepts or rejects each step is itself fragile. Absolute quality metrics neither indicate whether a step has improved the image~\cite{hu2022hierarchical} nor support fine-grained comparison among restored results~\cite{sheng2026fine}, and every wrongly accepted step both lowers fidelity and shifts later inputs away from the distributions on which downstream tools were trained~\cite{li2026hybrid,zhu2026opera}. The first two challenges are connected: restricting each step to one degradation type implicitly determines a restoration order, coupling pool construction with sequencing. All three challenges share one root: the true restoration objective is unmeasurable at inference time, so the agent must build a global trajectory from local decisions it can verify. Each local decision can fail in two ways: the right tool is absent from the candidate set, or a candidate is wrongly accepted.

To address these challenges, we present EORestore-Agent, a restoration agent that restores remote sensing images with composite degradations through a traceable trajectory of reference-free, verifiable per-step decisions. Our contributions are as follows:
\begin{itemize}
\item \textit{Multi-label degradation perception with a merged candidate pool.} A vision-language model (VLM) fine-tuned on remote sensing degradations re-perceives all residual atmospheric and sensor degradations at every step, and the tools of every type that still admits repair compete in one candidate pool of up to 17 tools, so the restoration order emerges from the evolving image rather than from a fixed plan. Prior agents either search the tools of one predicted type per step~\cite{chen2024restoreagent,zhu2025intelligent} or query each detected type separately~\cite{zhou2025q}. Compared with single-label perception, the merged pool raises PSNR by 0.37 to 0.57~dB when four or more degradations are present.
\item \textit{Relative quality scoring with a three-metric gate.} Because satellite acquisitions provide no clean reference, we make every restoration step verifiable by predicting its fidelity change. A relative quality scorer, learned from synthetic remote sensing degradation chains with known clean references, regresses the changes in PSNR, SSIM, and LPIPS from the input image to each candidate, instead of ranking candidates by absolute no-reference scores designed for natural images~\cite{zhou2025q}. The three predictions form a gate: a step is committed only when PSNR is predicted to improve and no metric is predicted to degrade, and because the unmodified input is always a candidate, the agent otherwise abstains~\cite{feng2024don} and keeps the current image. Replacing the learned scorer with differences of no-reference scores lowers PSNR by 1.1 to 4.6~dB, and the gate with abstention lowers the harmful-step rate from 28--40\% under ranking alone to 21--28\%.
\item \textit{Paradigm-level benchmarking with a step-level diagnosis.} We benchmark three restoration paradigms (twelve retrained all-in-one models, four natural-image restoration systems, and EORestore-Agent) on Landsat-8 composites of up to six atmospheric and sensor degradations, locating where each paradigm breaks down. All-in-one models trained on single degradations converge to a narrow band as degradations accumulate, with their PSNR range shrinking from 6.48~dB at one degradation to 1.12~dB at six, and natural-image agents evaluated zero-shot fall below the degraded input in PSNR for 17 of the 18 agent--$N$ pairs. At the step level, the remaining harmful steps of our agent trace to scorer overestimation near the acceptance threshold and, when only one degradation is present, to perception false positives.
\end{itemize}

\section{Related Work}
\label{sec:related}

\subsection{Unified and Composite Remote Sensing Image Restoration}
\label{sec:related_restoration}

Remote sensing image restoration has developed specialized models for atmospheric, sensor, and sampling degradations~\cite{rasti2021image,kabir2020vicarious,zhou2026remote}. These include generative and attention-based networks~\cite{spagan2020,zheng2020single,msda2022,acacrnet2024} and diffusion models~\cite{zou2024diffcr,sui2024diffusion} for cloud removal, physical scattering models~\cite{he2010single,guo2020landsat} and transformers~\cite{dehazeformer2023,psmbnet2023,rsdformer2023,aunet2024} for dehazing, and implicit image representations for continuous super-resolution, as in FunSR~\cite{funsr}. Together, these task-specific models provide a toolbox for restoring different aspects of a degraded image.

All-in-one (AIO) restoration trains a single network for multiple degradation types and has been surveyed recently~\cite{jiang2025survey}. AirNet~\cite{li2022all} learns degradation representations through contrastive learning, and PromptIR~\cite{potlapalli2023promptir} conditions a shared network on learned prompts. In remote sensing, Ada4DIR~\cite{lihe2025ada4dir} incorporates prompts derived from physical degradation models into a unified transformer. PhyDAE~\cite{dong2026phydae} extracts degradation cues through manifold and frequency analysis to guide physics-aware experts. CoRE-UIR~\cite{zhang2026core} adapts CLIP features into degradation priors and combines common and residual experts, with training that covers both single and compound degradations. TGPNet~\cite{huang2025task} modulates a shared decoder with learnable task embeddings and shows that unseen composites can be processed through sequential task prompting. PromptHSI~\cite{cheng2026prompthsi} extends vision-language guidance to hyperspectral restoration.

Composite degradation has also been studied with dedicated models. For natural images, dual-level prototype learning~\cite{wang2025dual} represents composites through degradation and factor prototypes, and OneRestore~\cite{guo2024onerestore} conditions a transformer on a scene descriptor encoding which degradations are present. In remote sensing, Wang and Zhang~\cite{wang2024unified} couple dehazing and super-resolution through saliency-guided interaction learning, and HDI-PRNet~\cite{feng2025progressive} unfolds a high-order degradation model into progressive denoising, deblurring, and super-resolution stages. These methods address multiple degradations through shared representations, task conditioning, or coupled restoration stages. Our work builds on pretrained restoration models as frozen tools and studies how to select and compose their operations for each observed image, including when to stop.

\subsection{Agent-Based Image Restoration}
\label{sec:related_agents}

Restoration agents determine which operation to apply next and use the resulting image to guide subsequent decisions. RL-Restore~\cite{yu2018crafting} established an early formulation of this process, training a reinforcement learning policy to select a small CNN from a toolbox or stop when restoring mixed distortions.

Language models support explicit planning and revision of restoration tasks. RestoreAgent~\cite{chen2024restoreagent} fine-tunes a multimodal language model to identify degradations, order the tasks, and choose a model for each. AgenticIR~\cite{zhu2025intelligent} adds reflection and rescheduling, using experience collected through self-exploration to support planning. MAIR~\cite{jiang2026multi} coordinates several agents over frozen tools under a prior on the order in which real degradations arise, while HybridAgent~\cite{li2026hybrid} combines single-task and mixed-distortion restoration modes. Such orchestration also supports applications with specific restoration requirements, including adverse-weather driving scenes in JarvisIR~\cite{lin2025jarvisir}.

Feedback from restoration outputs provides another basis for deciding the next step. Q-Agent~\cite{zhou2025q} queries each degradation type separately, evaluates candidate restorations with an aggregate of no-reference quality measures, and greedily selects the operation with the highest score. It stops when no operation is estimated to improve quality or all detected degradations have been addressed. Feedback also supports reflection and rollback in 4KAgent~\cite{zuo20254kagent}, while RAR~\cite{chen2026restore} couples restoration and assessment in an iterative latent-domain model.

Quality feedback can also train the policy that selects restoration actions. OPERA~\cite{zhu2026opera} learns complete tool sequences from a quality reward and finds that useful plans include tools outside the detected degradation set. DiTTo~\cite{choi2026ditto} aligns the policy to restoration order. TIR-Agent~\cite{jia2026tir} trains tool-calling policies with an adaptive multi-metric reward, and Restore-R1~\cite{lu2025restorer1} uses the per-step quality change from a multimodal model as its policy reward. These approaches incorporate restoration outcomes into the learning of action sequences.

In remote sensing, RIR-Agent~\cite{liu2026riragent} applies an agent framework with multimodal perception and tool execution to composite degradation restoration. Across these works, restoration agents typically restrict each step to the tools of one predicted degradation type~\cite{chen2024restoreagent,zhu2025intelligent}, and useful tools for co-occurring degradations can be excluded~\cite{zhu2026opera}. Our framework merges the tool pools of all detected types and accepts a step only when predicted fidelity changes are non-negative across multiple metrics.

\subsection{Quality Assessment for Restoration Decisions}
\label{sec:related_quality}

Restoration is evaluated with full-reference metrics when a clean image is available. SSIM~\cite{wang2004image} measures structural similarity, and LPIPS~\cite{zhang2018unreasonable} measures distance in deep feature space. Without a reference, no-reference image quality assessment estimates quality from the image alone. NIQE~\cite{niqe2013} measures departures from natural image statistics without fitting subjective ratings. MUSIQ~\cite{musiq2021} uses a transformer to aggregate quality information across image scales, and the no-reference variant of TOPIQ~\cite{chen2024topiq} guides distortion assessment with semantic features through attention across scales. Such no-reference scores can rank restored candidates and estimate improvement by comparison with the input, as demonstrated in quality-guided restoration agents~\cite{zhou2025q}.

Assessment tailored to restoration further models quality differences between closely related images. Hu et al.~\cite{hu2022hierarchical} learn hierarchical perceptual discrepancies directly from degraded and restored image pairs. Sheng et al.~\cite{sheng2026fine} introduce fine-grained preference annotations and combine quality regression with ranking to distinguish closely related restoration outputs. Quality signals have also served as feedback in restoration agents. RL-Restore~\cite{yu2018crafting} rewards each step proportional to quality improvement, and Restore-R1~\cite{lu2025restorer1} rewards its policy by the per-step change in a multimodal quality score.

Our scorer extends this perspective to optical remote sensing by predicting separate changes in PSNR, SSIM, and LPIPS from an input-candidate pair. Clean references supply the training targets but are unnecessary at inference. The predictions support ranking and a joint acceptance rule, with the current image retained when no eligible candidate has a positive predicted PSNR gain.

\section{Method}
\label{sec:method}

\subsection{Problem Formulation}
\label{sec:method_formulation}

Let $\mathbf{x}_{\mathrm{gt}}$ denote a clean remote sensing image. A compositely degraded observation is modeled as the result of $N$ degradation processes applied sequentially:
\begin{equation}
  \mathbf{x}_0 = D_N \circ D_{N-1} \circ \cdots \circ D_1 (\mathbf{x}_{\mathrm{gt}})
\label{eq:degradation}
\end{equation}
where each $D_i$ belongs to one of six degradation types $\mathcal{D} = \{d_1, \ldots, d_6\}$, namely cloud, haze, noise, blur, low light, and low resolution. Both the number of degradations $N$ and their types are unknown at inference time. Each type $d \in \mathcal{D}$ is associated with a pool of pretrained restoration tools $\mathcal{T}_d$, and the full library is $\mathcal{T} = \bigcup_{d \in \mathcal{D}} \mathcal{T}_d$.

Restoration proceeds as a sequence of actions $\mathbf{x}_{t+1} = a_t(\mathbf{x}_t)$, $t = 0, 1, \ldots$, with $a_t \in \mathcal{T} \cup \{\mathrm{id}\}$, where $\mathrm{id}$ leaves the image unchanged. The global objective for the action sequence to approximate is:
\begin{equation}
  \min_{a_0, \ldots, a_{K-1}} \, \rho\big( (a_{K-1} \circ \cdots \circ a_0)(\mathbf{x}_0), \, \mathbf{x}_{\mathrm{gt}} \big)
\label{eq:global}
\end{equation}
where $\rho(\cdot, \cdot)$ is a full-reference distortion measure and $K$ bounds the trajectory length. A global optimum of \eqref{eq:global} exists, but no agent can search for it directly since the reference $\mathbf{x}_{\mathrm{gt}}$ is unavailable at inference time.

\subsection{Overview of EORestore-Agent}
\label{sec:method_overview}

EORestore-Agent turns the global objective into per-step decisions. The total fidelity gain in \eqref{eq:global} is the sum of the per-step changes $\rho(\mathbf{x}_t, \mathbf{x}_{\mathrm{gt}}) - \rho(\mathbf{x}_{t+1}, \mathbf{x}_{\mathrm{gt}})$, so the objective improves monotonically along a trajectory whose every step improves fidelity. To assess the improvement accurately, a learned relative quality scorer $f_{\theta}$ is designed to output $\hat{\Delta}_{P}$, $\hat{\Delta}_{S}$, and $\hat{\Delta}_{L}$, the predicted changes in PSNR, SSIM, and LPIPS between the input image and a candidate, all oriented so that positive values indicate improvement (Section~\ref{sec:method_scoring}).

At each step $t$, the agent assembles a merged pool $\mathcal{U}_t$ of candidate tools from the degradation types it detects in the input image (Sections~\ref{sec:method_perception} and \ref{sec:method_pool}) and selects the action
\begin{equation}
\begin{aligned}
a_t^{\star}
&= \arg\max_{a \in \mathcal{U}_t \cup \{\mathrm{id}\}}
\hat{\Delta}_{P}\big(\mathbf{x}_t, a(\mathbf{x}_t)\big) \\
&\quad \text{s.t.}\quad
\hat{\Delta}_{P} \geq 0,\quad
\hat{\Delta}_{S} \geq 0,\quad
\hat{\Delta}_{L} \geq 0
\end{aligned}
\label{eq:rule}
\end{equation}

The constraints admit only actions whose predicted deltas are all non-negative, so no accepted action is predicted to trade one quality dimension against another. Because $\mathrm{id}$ has all three deltas equal to zero by definition, the agent can always abstain, and it commits to $a_t^{\star}$ only when the predicted PSNR gain is strictly positive.

\begin{figure*}[t]
\centering
\includegraphics[width=\textwidth]{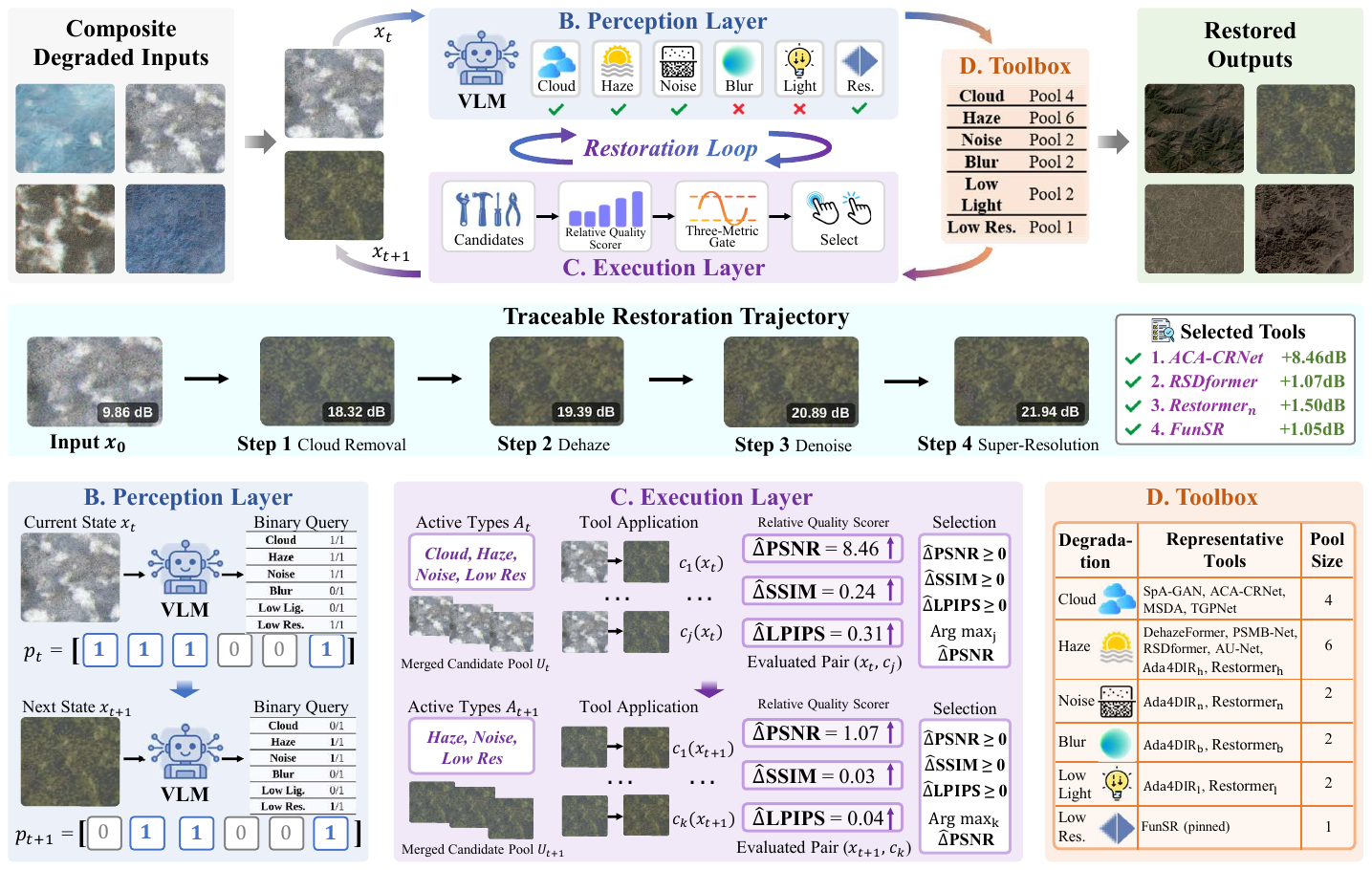}
\caption{Overview of EORestore-Agent. Top: the iterative restoration loop with perception, merged candidate pool, relative quality scoring, and three-metric gating. Middle: an example trajectory on a tile with four degradation types, showing the selected tool and PSNR at each step. Bottom: detailed view of the perception layer, execution layer, and toolbox.}
\label{fig:pipeline}
\end{figure*}

EORestore-Agent replaces the global search of \eqref{eq:global} with \eqref{eq:rule}, making one decision per step, and how closely it approaches the global optimum depends on the two components: the merged candidate pool $\mathcal{U}_t$ and the predicted deltas $\hat{\Delta}_{P}$, $\hat{\Delta}_{S}$, and $\hat{\Delta}_{L}$. The remainder of this section builds one mechanism for each component. Fig.~\ref{fig:pipeline} illustrates the overall pipeline. The agent iterates perception, pool construction, tool execution, and scoring-based selection until no degradation type remains active or a maximum of $K$ iterations is reached. A per-type repair budget $n$ and a closure mechanism (Section~\ref{sec:method_scoring}) guarantee termination. The complete procedure is summarized in Algorithm~\ref{alg:pipeline} in Section~\ref{sec:method_algorithm}.

\subsection{Multi-Label Degradation Perception}
\label{sec:method_perception}

At each iteration, the agent must determine which degradation types are present in the input image. We formulate this as six independent binary visual question answering (VQA) queries, one per degradation type, posed to a fine-tuned vision-language model (VLM). Writing $\Phi(\mathbf{x}_t, q_d) \in \{0, 1\}$ for the VLM's answer to the query $q_d$ about type $d$, the perceived set is
\begin{equation}
  \mathcal{P}_t = \{\, d \in \mathcal{D} : \Phi(\mathbf{x}_t, q_d) = 1 \,\}
\label{eq:perception}
\end{equation}
The prompt templates are provided in Appendix~\ref{app:prompts}.

We use Qwen3.5-9B as the backbone and finetune it with low-rank adaptation (LoRA, rank 32) on a training set of synthetically degraded remote sensing images with known degradation labels. The visual encoder is frozen during finetuning, and the same prompt template is used for both training and inference. Training details are provided in Section~\ref{sec:experiments}.

The multi-label output means that no co-occurring degradation is discarded in favor of a single dominant label, so every detected type can contribute tools to the pool constructed in Section~\ref{sec:method_pool}. Re-executing the perception at every iteration adapts the pool to the evolving state of the image.

\subsection{Merged Candidate Pool and Execution}
\label{sec:method_pool}

Given the perceived set $\mathcal{P}_t$, the agent gathers candidate tools from every detected type that still admits repair. Writing $\mathcal{A}_t \subseteq \mathcal{P}_t$ for these active types, where a type leaves $\mathcal{A}_t$ once its repair budget $n$ is exhausted or it has been closed by input fallback (Section~\ref{sec:method_scoring}, Algorithm~\ref{alg:pipeline}), the candidate pool merges the tool pools of all active types:
\begin{equation}
  \mathcal{U}_t = \bigcup_{d \in \mathcal{A}_t} \mathcal{T}_d
\label{eq:pool}
\end{equation}
The pool is deliberately not restricted to the tools of any single type. When multiple degradations coexist, a tool nominally designed for one type may also improve another, and merging keeps such cross-type options available for the subsequent scoring. Table~\ref{tab:toolbox} lists the tools and their associated types.

\begin{table}[t]
\caption{Restoration Toolbox}
\label{tab:toolbox}
\centering
\footnotesize
\renewcommand{\arraystretch}{1.25}
\begin{tabularx}{\columnwidth}{@{}l>{\raggedright\arraybackslash}Xc@{}}
\toprule
Type & Tools & Size \\
\midrule
Cloud & SpA-GAN~\cite{spagan2020}, ACA-CRNet~\cite{acacrnet2024}, \mbox{MSDA-CR~\cite{msda2022,sui2024diffusion}}, TGPNet~\cite{huang2025task} & 4 \\
Haze & DehazeFormer~\cite{dehazeformer2023}, PSMB-Net~\cite{psmbnet2023}, RSDformer~\cite{rsdformer2023}, AU-Net~\cite{aunet2024}, Ada4DIR$_\text{h}$, Restormer$_\text{h}$ & 6 \\
Noise & Ada4DIR$_\text{n}$, Restormer$_\text{n}$ & 2 \\
Blur & Ada4DIR$_\text{b}$, Restormer$_\text{b}$ & 2 \\
Low light & Ada4DIR$_\text{l}$, Restormer$_\text{l}$ & 2 \\
Low resolution & FunSR~\cite{funsr} & 1 \\
\bottomrule
\end{tabularx}
\par\vspace{3pt}
{\scriptsize\raggedright\noindent Subscripts h, n, b, and l denote Ada4DIR~\cite{lihe2025ada4dir} or Restormer~\cite{zamir2022restormer} models trained from scratch on a single degradation type (haze, noise, blur, or low light). At each step, the pools of all active types are merged and scored by the relative quality scorer.\par}
\end{table}

Every tool in the merged pool is applied to the input image, and the unmodified input image is appended as the no-action baseline:
\begin{equation}
  \mathcal{X}_t = \{\, a(\mathbf{x}_t) : a \in \mathcal{U}_t \,\} \cup \{\, \mathbf{x}_t \,\}
\label{eq:candidates}
\end{equation}
Together with the multi-label perception, the merged pool keeps every plausible action under consideration at each step. Ranking the candidates and deciding whether any of them should be accepted is the task of the scorer described next.

\subsection{Relative Quality Scoring and Three-Metric Gate}
\label{sec:method_scoring}

The candidate set $\mathcal{X}_t$ contains outputs of tools built for different degradation types, so selecting among them requires a common quality scale. We use the quality change from the input image to each candidate, which directly answers the decision the agent faces at every step, namely whether applying a tool leaves the image better than not applying it.

For each candidate $\mathbf{c} \in \mathcal{X}_t$, the scorer takes the input image $\mathbf{x}_t$ and the candidate as a pair and outputs three predicted quality changes:
\begin{equation}
  f_{\theta}(\mathbf{x}_t, \mathbf{c}) = \big(\, \hat{\Delta}_{P}, \ \hat{\Delta}_{S}, \ \hat{\Delta}_{L} \,\big)
\label{eq:scorer}
\end{equation}
where $\hat{\Delta}_{P}$, $\hat{\Delta}_{S}$, and $\hat{\Delta}_{L}$ estimate the change in PSNR, SSIM, and LPIPS that accepting $\mathbf{c}$ would bring over keeping $\mathbf{x}_t$. The sign of the LPIPS estimate is flipped so that positive values indicate improvement on all three metrics.

The scorer uses a frozen DINOv3 ViT-L/16 pretrained on satellite imagery (SAT-493M)~\cite{simeoni2025dinov3} to extract semantic features from both images, complemented by 16 handcrafted statistics per spatial region covering brightness, color, gradient, and frequency properties. Both feature branches are frozen and require no training. The difference between the candidate and input-image features is summarized into a fixed-length vector and passed through a lightweight two-layer multilayer perceptron (MLP) with three output heads, one per delta. To reduce prediction variance, $E$ independently trained members are ensembled and their per-metric predictions are averaged:
\begin{equation}
  \hat{\Delta}_{M}(\mathbf{x}_t, \mathbf{c}) = \frac{1}{E} \sum_{e=1}^{E} f_{\theta_e}^{M}(\mathbf{x}_t, \mathbf{c}), \quad M \in \{P, S, L\}
\label{eq:ensemble}
\end{equation}

For the input image included as its own candidate, the three deltas are defined as exactly zero, since no change has occurred. With this sign convention, the true deltas are positive for a tool that improves the image and negative for one that harms it. The scorer is trained to predict them.

Selection then implements rule~\eqref{eq:rule} over the scored candidates. A candidate is eligible only when all three predicted deltas are non-negative, a condition we refer to as the three-metric gate. The candidates that pass it form the eligible set $\mathcal{G}_t$. Among the eligible candidates, the one with the highest predicted PSNR improvement is accepted as the result of the current step. When no eligible candidate offers a strictly positive predicted PSNR gain, the agent abstains. 
Abstention is a model's deliberate choice to withhold an output that is likely to be wrong, putting reliability ahead of always responding~\cite{feng2024don}. In EORestore-Agent, abstention means applying no tool and retaining the input image of the current step, which we refer to as input fallback.
The active types are then permanently closed, which stops the agent from retrying types that its tools cannot improve. Algorithm~\ref{alg:pipeline} gives the complete procedure.

The scorer is trained on pairs of (input image, candidate output) constructed from synthetic degradation chains applied to remote sensing training images, where the clean reference of every chain is known. For a training pair $(\mathbf{x}, \mathbf{c})$, the supervision target is the ground-truth quality change measured against the reference:
\begin{equation}
\begin{aligned}
  \Delta_{P}(\mathbf{x}, \mathbf{c}) &= \mathrm{PSNR}(\mathbf{c}, \mathbf{x}_{\mathrm{gt}}) - \mathrm{PSNR}(\mathbf{x}, \mathbf{x}_{\mathrm{gt}}) \\
  \Delta_{S}(\mathbf{x}, \mathbf{c}) &= \mathrm{SSIM}(\mathbf{c}, \mathbf{x}_{\mathrm{gt}}) - \mathrm{SSIM}(\mathbf{x}, \mathbf{x}_{\mathrm{gt}}) \\
  \Delta_{L}(\mathbf{x}, \mathbf{c}) &= \mathrm{LPIPS}(\mathbf{x}, \mathbf{x}_{\mathrm{gt}}) - \mathrm{LPIPS}(\mathbf{c}, \mathbf{x}_{\mathrm{gt}})
\end{aligned}
\label{eq:supervision}
\end{equation}
Training details are provided in Section~\ref{sec:experiments}. By admitting only candidates that pass the three-metric gate and by allowing the agent to abstain, the selection limits the entry of harmful steps into the trajectory.

\subsection{Algorithm Summary}
\label{sec:method_algorithm}

Algorithm~\ref{alg:pipeline} assembles the components of Sections~\ref{sec:method_perception} to \ref{sec:method_scoring} into the complete restoration loop. Each iteration runs the perception \eqref{eq:perception}, builds the merged pool \eqref{eq:pool}, executes the candidates \eqref{eq:candidates}, and scores and selects among them as described in Section~\ref{sec:method_scoring}. The loop ends when no active type remains or after $K$ iterations. For readability, Algorithm~\ref{alg:pipeline} drops the step index $t$ and updates $\mathbf{x}$ in place.

\begin{algorithm}[t]
\caption{EORestore-Agent Iterative Restoration}
\label{alg:pipeline}
\begin{algorithmic}[1]
\REQUIRE degraded image $\mathbf{x}_0$, tool pools $\{\mathcal{T}_d\}_{d \in \mathcal{D}}$, repair budget $n$, max iterations $K$
\ENSURE restored image $\mathbf{x}$
\STATE $\mathbf{x} \leftarrow \mathbf{x}_0$
\STATE $r_d \leftarrow 0$ for every type $d \in \mathcal{D}$ \COMMENT{per-type repair counter}
\STATE $\mathcal{C} \leftarrow \emptyset$ \COMMENT{types closed by input fallback}
\FOR{$t = 0$ \TO $K-1$}
  \STATE $\mathcal{P} \leftarrow \{\, d \in \mathcal{D} : \Phi(\mathbf{x}, q_d) = 1 \,\}$ \COMMENT{Eq.~\eqref{eq:perception}}
  \STATE $\mathcal{A} \leftarrow \mathcal{P} \setminus \{\, d : r_d \geq n \,\} \setminus \mathcal{C}$ \COMMENT{within budget, not closed}
  \IF{$\mathcal{A} = \emptyset$}
    \STATE \textbf{break}
  \ENDIF
  \STATE $\mathcal{U} \leftarrow \bigcup_{d \in \mathcal{A}} \mathcal{T}_d$ \COMMENT{Eq.~\eqref{eq:pool}}
  \STATE $\mathcal{X} \leftarrow \{\, a(\mathbf{x}) : a \in \mathcal{U} \,\} \cup \{\, \mathbf{x} \,\}$ \COMMENT{Eq.~\eqref{eq:candidates}}
  \FOR{each candidate $\mathbf{c} \in \mathcal{X}$}
    \IF{$\mathbf{c} = \mathbf{x}$}
      \STATE $(\hat{\Delta}_{P}, \hat{\Delta}_{S}, \hat{\Delta}_{L}) \leftarrow (0, 0, 0)$
    \ELSE
      \STATE $(\hat{\Delta}_{P}, \hat{\Delta}_{S}, \hat{\Delta}_{L}) \leftarrow$ ensemble scorer \COMMENT{Eq.~\eqref{eq:ensemble}}
    \ENDIF
  \ENDFOR
  \STATE $\mathcal{G} \leftarrow \{\, \mathbf{c} \in \mathcal{X} : \hat{\Delta}_{P} \geq 0,\ \hat{\Delta}_{S} \geq 0,\ \hat{\Delta}_{L} \geq 0 \,\}$ \COMMENT{three-metric gate}
  \IF{$\max_{\mathbf{c} \in \mathcal{G}} \hat{\Delta}_{P} > 0$}
    \STATE $\mathbf{c}^{\star} \leftarrow \arg\max_{\mathbf{c} \in \mathcal{G}} \hat{\Delta}_{P}$ \COMMENT{best eligible candidate}
    \STATE $\mathbf{x} \leftarrow \mathbf{c}^{\star}$
    \STATE $r_{\mathrm{type}(\mathbf{c}^{\star})} \leftarrow r_{\mathrm{type}(\mathbf{c}^{\star})} + 1$ \COMMENT{type of the tool that produced $\mathbf{c}^{\star}$}
  \ELSE
    \STATE $\mathcal{C} \leftarrow \mathcal{C} \cup \mathcal{A}$ \COMMENT{input fallback, permanent closure}
  \ENDIF
\ENDFOR
\RETURN $\mathbf{x}$
\end{algorithmic}
\end{algorithm}

\section{Experiments}
\label{sec:experiments}

\subsection{Experimental Setup}
\label{sec:exp_setup}

We build a synthetic benchmark from the MDRS-Landsat dataset, which provides $512 \times 512$ RGB tiles from Landsat-8 optical imagery. The test set contains 270 clean tiles. Six degradation types are considered: cloud ($d_1$), haze ($d_2$), noise ($d_3$), blur ($d_4$), low light ($d_5$), and low resolution ($d_6$). For each number of degradations $N \in \{1, \ldots, 6\}$, we construct degradation chains by selecting subsets of these six types and applying them sequentially to a clean tile. Severity parameters are drawn independently at random for each degradation and each tile. The benchmark covers $N{=}1$ (single degradation) through $N{=}6$ (all six types), with 34 degradation chains distributed across the six values of $N$. At $N{=}6$, the four chains apply the same six types in different orders. Each combination is evaluated on all 270 test tiles. Table~\ref{tab:main} reports averages per $N$, and per-combination results are provided in Appendix~\ref{app:percombo}. The six degradation types are synthesized as follows:
\begin{enumerate}
\item Haze follows the wavelength-dependent atmospheric scattering model of RS-Haze~\cite{dehazeformer2023}, with airlight $A \sim \mathrm{Unif}(0.85, 0.95)$ and per-band transmission obtained from sampled transmission fields.
\item Cloud is generated by SatelliteCloudGenerator~\cite{rs15174138}, a physically based simulator that produces localized cloud masks with cast shadows for satellite imagery.
\item Blur applies convolution with a random anisotropic Gaussian kernel (support 3 to 13, $\sigma_b \sim \mathrm{Unif}(0.2, 4.0)$ pixels).
\item Noise is additive white Gaussian with $\sigma_n \sim \mathrm{Unif}(5, 25)$ on the 8-bit scale.
\item Low light applies a power-law transform $I = J^{\gamma}$ to the normalized input intensity $J$ with $\gamma \sim \mathrm{Unif}(1.2, 2.5)$.
\item Low resolution performs bicubic $4\times$ downsampling, reducing the tile to $128 \times 128$.
\end{enumerate}

We compare against three groups of baselines. The first group contains nine natural-image all-in-one (AIO) models: PromptIR~\cite{potlapalli2023promptir}, Restormer~\cite{zamir2022restormer}, NAFNet~\cite{chen2022simple}, DGUNet~\cite{mou2022deep}, IDR~\cite{zhang2023ingredient}, AirNet~\cite{li2022all}, and three backbones (SRResNet, Uformer, Restormer) trained with the sequential and adaptive prompt learning of MiOIR~\cite{kong2024towards}, denoted SRResNet-S+AP, Uformer-S+AP, and Restormer-S+AP. These three models extract task-specific information from the degraded input itself, requiring no explicit task identifier. The second group contains three remote-sensing AIO models: PhyDAE~\cite{dong2026phydae}, Ada4DIR~\cite{lihe2025ada4dir}, and CoRE-UIR~\cite{zhang2026core}. All AIO models in these two groups are retrained on the MDRS-Landsat training split, which covers haze, noise, blur, and low light. Cloud and low resolution are therefore unseen for every AIO baseline. The third group contains four natural-image restoration systems, evaluated zero-shot with their released weights: AutoDIR~\cite{jiang2024autodir} and the three restoration agents AgenticIR~\cite{zhu2025intelligent}, 4KAgent~\cite{zuo20254kagent}, and JarvisIR~\cite{lin2025jarvisir}. AutoDIR is designed as a foundation model that detects unknown degradations and restores them step by step, and its authors evaluate it on unseen datasets without retraining. We follow the same protocol. This group shows how natural-image multi-step restoration systems transfer to remote sensing imagery. Together with EORestore-Agent, the three groups span three restoration paradigms: all-in-one models (the first two groups), natural-image restoration systems (the third group), and our agent. Restoration quality is measured by PSNR (dB, $\uparrow$), SSIM ($\uparrow$), and LPIPS ($\downarrow$), all computed between the restored output and the clean reference.

The per-type tools Ada4DIR$_\text{h,n,b,l}$ and Restormer$_\text{h,n,b,l}$ (Table~\ref{tab:toolbox}) keep the architectures of Ada4DIR and Restormer but are trained from scratch, one model per degradation type, on single-degradation images from the MDRS-Landsat training split. The training split does not overlap with the 270 test tiles. These tools are therefore distinct from the AIO baselines of the same name in Table~\ref{tab:main}. The pool sizes in Table~\ref{tab:toolbox} reflect model availability: cloud removal and dehazing have multiple published remote sensing models with open-source weights, whereas denoising, deblurring, and low-light enhancement lack suitable task-specific models for remote sensing, so we trained the two strongest AIO architectures as single-task specialists from scratch to fill these pools. For super-resolution, FunSR consistently outperformed other candidates in preliminary evaluation and is retained as the sole tool.

The multi-label perception (Section~\ref{sec:method_perception}) finetunes Qwen3.5-9B with LoRA (rank 32, $\alpha = 64$) on synthetically degraded images from the MDRS-Landsat training split. Each degraded image is paired with binary VQA queries for each degradation type, producing approximately 50{,}000 training samples. The visual encoder is frozen, and the same prompt template is used for training and inference. Training runs for one epoch with a cosine learning rate schedule (peak rate $10^{-4}$, warmup ratio 0.03), and an effective batch size of 128.

The relative quality scorer (Section~\ref{sec:method_scoring}) is trained on 5{,}030 tiles from the training split, with 100 tiles held out for validation. For each tile, we sample degradation chains as random full-length permutations of the six types. Because each chain is synthesized by adding one degradation at a time, every intermediate image is available at no extra cost and serves as a training input for the scorer. Each intermediate image is paired with its restoration candidates and labeled by the change in PSNR, SSIM, and LPIPS against the clean reference. The frozen DINOv3 ViT-L/16 backbone~\cite{simeoni2025dinov3} encodes patch tokens from three depths, pooled onto a coarse spatial grid alongside low-level regional statistics. The prediction head is trained with a robust regression loss and a within-image ranking term, using AdamW (learning rate $10^{-3}$, weight decay $10^{-2}$) for 30 epochs over 397{,}593 input-candidate pairs. We train $E{=}7$ models differing only in random seed and average their predictions at inference. EORestore-Agent operates with a per-type repair budget of $n{=}3$ and a maximum of $K{=}15$ iterations per tile.

\subsection{Paradigm-Level Benchmarking}
\label{sec:exp_main}

\begin{table*}[t]
\caption{Comparison on the synthetic benchmark. PSNR (dB, $\uparrow$), SSIM ($\uparrow$), and LPIPS ($\downarrow$) averaged over all degradation combinations at each $N$. Best and second best among restoration methods are in bold and underlined. The degraded input is not ranked. Ties share the mark.}
\label{tab:main}
\centering
\footnotesize
\renewcommand{\arraystretch}{0.95}
\resizebox{\textwidth}{!}{\begin{tabular}{@{}lrrrrrrrrrrrrrrrrrr@{}}
\toprule
 & \multicolumn{18}{c}{Number of composited degradations} \\
\cmidrule(l){2-19}
 & \multicolumn{3}{c}{1} & \multicolumn{3}{c}{2} & \multicolumn{3}{c}{3} & \multicolumn{3}{c}{4} & \multicolumn{3}{c}{5} & \multicolumn{3}{c}{6} \\
\cmidrule(lr){2-4} \cmidrule(lr){5-7} \cmidrule(lr){8-10} \cmidrule(lr){11-13} \cmidrule(lr){14-16} \cmidrule(l){17-19}
Method & PSNR & SSIM & LPIPS & PSNR & SSIM & LPIPS & PSNR & SSIM & LPIPS & PSNR & SSIM & LPIPS & PSNR & SSIM & LPIPS & PSNR & SSIM & LPIPS \\
\midrule
Degraded input & 22.93 & 0.706 & 0.377 & 18.84 & 0.562 & 0.586 & 16.09 & 0.476 & 0.770 & 14.24 & 0.358 & 0.874 & 14.66 & 0.362 & 0.915 & 14.45 & 0.412 & 0.877 \\
\midrule
\multicolumn{19}{@{}l}{\emph{Natural-image all-in-one}} \\
PromptIR~\cite{potlapalli2023promptir} & 31.82 & 0.887 & 0.193 & 19.40 & 0.709 & 0.411 & 17.29 & 0.597 & 0.597 & 16.29 & 0.537 & 0.696 & 15.82 & 0.482 & 0.774 & 15.61 & 0.478 & 0.809 \\
Restormer~\cite{zamir2022restormer} & 32.64 & 0.889 & 0.196 & \underline{22.81} & \underline{0.744} & 0.392 & \underline{19.04} & \underline{0.629} & 0.585 & \underline{18.13} & \textbf{0.574} & \underline{0.661} & \underline{17.12} & \textbf{0.530} & 0.730 & \underline{16.35} & \textbf{0.513} & 0.775 \\
NAFNet~\cite{chen2022simple} & \underline{34.27} & \underline{0.893} & 0.188 & 20.60 & 0.727 & 0.405 & 17.27 & 0.612 & 0.596 & 16.24 & 0.540 & 0.686 & 15.54 & 0.487 & 0.770 & 15.59 & 0.485 & 0.805 \\
DGUNet~\cite{mou2022deep} & 31.73 & 0.886 & 0.192 & 19.85 & 0.702 & 0.411 & 17.32 & 0.580 & 0.600 & 17.00 & 0.549 & 0.697 & 16.08 & 0.485 & 0.780 & 15.52 & 0.478 & 0.808 \\
IDR~\cite{zhang2023ingredient} & 30.15 & 0.878 & 0.203 & 20.35 & 0.695 & 0.436 & 18.10 & 0.578 & 0.635 & 17.50 & 0.507 & 0.731 & 16.29 & 0.433 & 0.804 & 15.78 & 0.443 & 0.814 \\
SRResNet-S+AP~\cite{kong2024towards} & 31.64 & 0.889 & 0.192 & 20.22 & 0.716 & 0.419 & 17.25 & 0.598 & 0.603 & 16.15 & 0.524 & 0.689 & 15.52 & 0.465 & 0.776 & 15.23 & 0.457 & 0.814 \\
Uformer-S+AP~\cite{kong2024towards} & 33.45 & \textbf{0.895} & 0.183 & 20.63 & 0.729 & 0.392 & 17.72 & 0.611 & 0.585 & 16.57 & 0.545 & 0.682 & 15.96 & 0.484 & 0.771 & 15.61 & 0.488 & 0.804 \\
Restormer-S+AP~\cite{kong2024towards} & 33.42 & 0.892 & 0.180 & 20.71 & 0.731 & 0.387 & 17.56 & 0.618 & 0.578 & 16.53 & \underline{0.554} & 0.673 & 16.03 & 0.492 & 0.760 & 15.71 & 0.485 & 0.799 \\
AirNet~\cite{li2022all} & 28.13 & 0.865 & 0.237 & 21.11 & 0.706 & 0.444 & 18.57 & 0.588 & 0.641 & 17.38 & 0.504 & 0.734 & 16.42 & 0.438 & 0.807 & 16.03 & 0.445 & 0.814 \\
\midrule
\multicolumn{19}{@{}l}{\emph{Remote-sensing all-in-one}} \\
PhyDAE~\cite{dong2026phydae} & 32.28 & 0.885 & \underline{0.171} & 19.85 & 0.720 & \underline{0.380} & 17.06 & 0.607 & \underline{0.565} & 16.08 & 0.547 & 0.672 & 15.54 & 0.493 & 0.761 & 15.46 & 0.487 & 0.811 \\
Ada4DIR~\cite{lihe2025ada4dir} & 33.54 & \underline{0.893} & 0.186 & 20.88 & 0.728 & 0.413 & 17.44 & 0.612 & 0.602 & 16.04 & 0.528 & 0.709 & 15.54 & 0.470 & 0.784 & 15.28 & 0.475 & 0.813 \\
CoRE-UIR~\cite{zhang2026core} & \textbf{34.61} & 0.887 & \textbf{0.160} & 20.74 & 0.713 & 0.390 & 17.06 & 0.581 & 0.584 & 16.05 & 0.509 & 0.685 & 15.80 & 0.482 & 0.769 & 15.62 & 0.485 & 0.815 \\
\midrule
\multicolumn{19}{@{}l}{\emph{Natural-image systems (zero-shot)}} \\
AutoDIR~\cite{jiang2024autodir} & 18.63 & 0.659 & 0.374 & 14.71 & 0.535 & 0.571 & 13.13 & 0.431 & 0.753 & 12.21 & 0.399 & 0.818 & 12.17 & 0.369 & 0.904 & 12.27 & 0.394 & 0.923 \\
AgenticIR~\cite{zhu2025intelligent} & 16.84 & 0.556 & 0.418 & 16.13 & 0.497 & 0.526 & 15.03 & 0.435 & 0.619 & 14.61 & 0.350 & 0.706 & 13.95 & 0.335 & 0.723 & 13.48 & 0.316 & \underline{0.706} \\
4KAgent~\cite{zuo20254kagent} & 14.47 & 0.425 & 0.503 & 14.33 & 0.454 & 0.521 & 14.10 & 0.408 & 0.573 & 13.47 & 0.369 & \underline{0.661} & 13.68 & 0.378 & \textbf{0.671} & 13.69 & 0.372 & \textbf{0.671} \\
JarvisIR~\cite{lin2025jarvisir} & 15.14 & 0.529 & 0.486 & 13.94 & 0.507 & 0.573 & 12.24 & 0.455 & 0.682 & 11.45 & 0.431 & 0.729 & 10.57 & 0.413 & 0.790 & 10.38 & 0.402 & 0.810 \\
\midrule
\multicolumn{19}{@{}l}{\emph{EORestore-Agent}} \\
Ours & 33.66 & 0.882 & 0.182 & \textbf{26.05} & \textbf{0.771} & \textbf{0.360} & \textbf{22.07} & \textbf{0.650} & \textbf{0.540} & \textbf{20.56} & 0.550 & \textbf{0.649} & \textbf{19.44} & \underline{0.509} & \underline{0.718} & \textbf{18.98} & \underline{0.500} & 0.745 \\
\bottomrule\end{tabular}}
\end{table*}

\begin{figure}[t]
\centering
\includegraphics[width=\columnwidth]{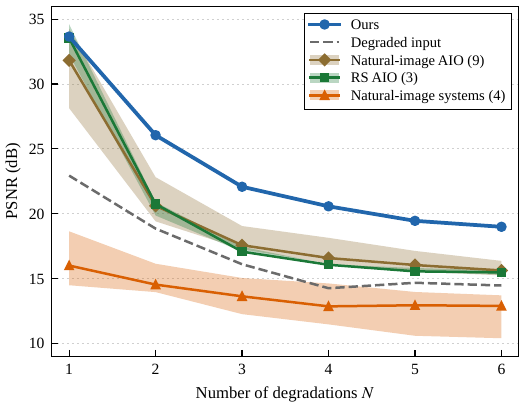}
\caption{PSNR versus the number of composited degradations $N$, plotted from Table~\ref{tab:main}. Each band spans the minimum and maximum of a baseline group, and the line with markers is the group median. The legend gives the number of models in each group.}
\label{fig:psnr_vs_n}
\end{figure}

Table~\ref{tab:main} compares all methods across $N{=}1$ to $6$. At $N{=}1$, each test sample contains a single degradation, and the AIO baselines trained for single-degradation restoration perform best: CoRE-UIR leads at 34.61 dB, 0.95 dB above EORestore-Agent (33.66 dB). This gap reflects the AIO advantage on the four degradation types that their training covers (blur, noise, haze, low light). On cloud, which falls outside the training data of every AIO baseline, EORestore-Agent leads the best baseline by 4.88 dB (23.36 vs.\ 18.48). From $N{=}2$ onward, EORestore-Agent leads all baselines in PSNR at every level. The margin over the second-best method (Restormer) ranges from 2.3 to 3.2 dB. Fig.~\ref{fig:psnr_vs_n} visualizes these trends.

As $N$ increases, the AIO models converge to a narrow performance band regardless of architecture. Among the twelve retrained AIOs, the PSNR range spans 6.48 dB at $N{=}1$ (from AirNet at 28.13 to CoRE-UIR at 34.61). By $N{=}6$, this range collapses to 1.12 dB (from SRResNet-S+AP at 15.23 to Restormer at 16.35). The three remote-sensing AIOs (PhyDAE, Ada4DIR, CoRE-UIR) converge to within 0.34 dB of each other at $N{=}6$. A similar narrow band has been reported for AIO models on unseen composite degradations of remote sensing images~\cite{huang2025task}. Restormer marks the upper edge of this band. On the six single degradations it ranks between fifth and tenth among the twelve AIOs, yet it is the best AIO on 18 of the 28 composite combinations (Appendix~\ref{app:percombo}). It is retrained on the same data as the other AIOs. We therefore treat it as the strongest baseline in the comparisons that follow. Single-model restoration approaches share a performance ceiling on composite degradation that architecture choice alone cannot lift.

The three natural-image restoration agents (AgenticIR, 4KAgent, JarvisIR) produce outputs below the degraded input in PSNR at every $N$, with one exception (AgenticIR at $N{=}4$, 14.61 vs.\ 14.24 dB). AutoDIR, also evaluated zero-shot, stays below the degraded input at every $N$ (18.63 vs.\ 22.93 dB at $N{=}1$). At $N{=}1$, the best agent (AgenticIR, 16.84 dB) falls 6.09 dB below the degraded input (22.93 dB). At $N{=}6$, JarvisIR drops to 10.38 dB, 4.07 dB below the degraded input (14.45 dB). The restoration tools and quality criteria of these agents were built for natural images, and they mostly lower the PSNR of remote sensing images.

\begin{figure*}[t]
\centering
\includegraphics[width=\textwidth]{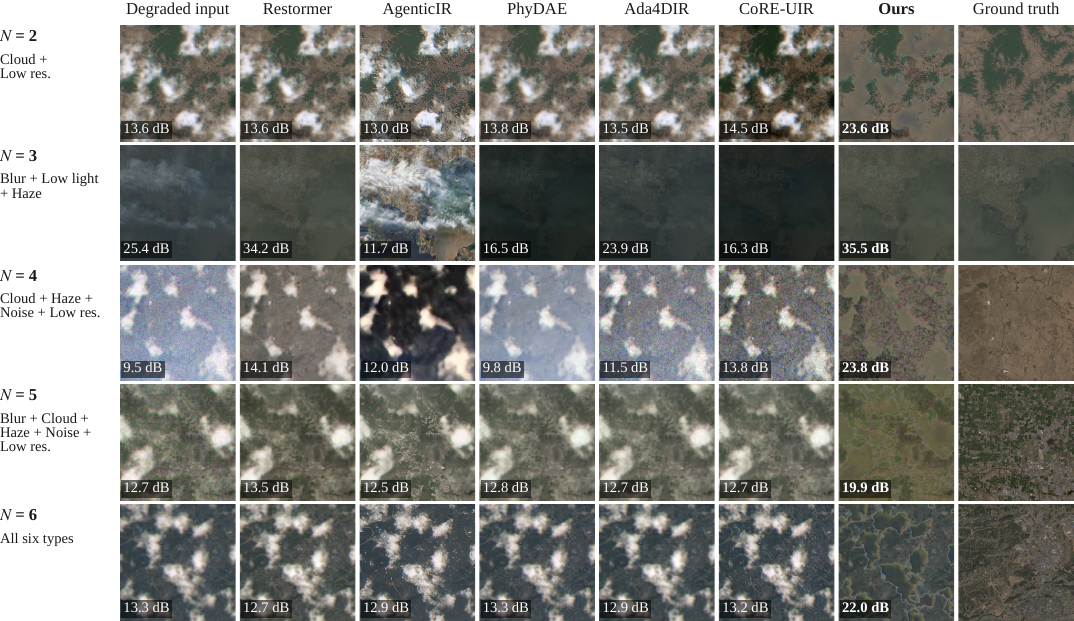}
\caption{Qualitative comparison on the synthetic benchmark, one case per $N$ from 2 to 6. Row labels give the degradation combination. The PSNR of each output is shown at the bottom left, with the best value of each row in bold.}
\label{fig:qualitative}
\end{figure*}

Per-combination results (Appendix~\ref{app:percombo}) sharpen these findings. The lead has two sources. Cloud and low resolution lie outside the training data of every AIO baseline. Averaged over the combinations that contain cloud at each $N$, EORestore-Agent leads the best baseline by 2.6 to 5.9 dB. Averaged over the combinations without cloud, the lead is 0.4 to 2.0 dB at $N \geq 2$. The smaller margin is within the range that restoration agents report over AIO models on natural images~\cite{zhu2025intelligent,chen2024restoreagent}. The larger margin shows the value of a toolbox that covers new degradation types without retraining a single model. Cloud combined with low resolution is the hardest case for the AIO models. At $N{=}2$ on cloud+low resolution, every AIO scores below the degraded input (best: AirNet at 16.56, degraded: 16.63), while EORestore-Agent reaches 23.08 dB. The agents score lower still, with JarvisIR at 13.69 dB on the same combination. Fig.~\ref{fig:qualitative} provides visual evidence. On the $N{=}2$, $N{=}5$, and $N{=}6$ cases, the three remote-sensing AIO models score within 1 dB of each other and of the degraded input despite differing architectures, consistent with the quantitative convergence above. AgenticIR introduces visible artifacts, and its output falls below the degraded input in four of the five cases.

\subsection{Comparison on Seen Degradation Combinations}
\label{sec:exp_seen}

\begin{table}[t]
\caption{Comparison on combinations composed entirely of seen degradation types.
Only the four types covered by AIO training (haze, noise, blur, low light) are included. Restormer is the strongest baseline at $N \geq 2$ in Table~\ref{tab:main}. The $N{=}4$ column contains a single combination. Best in bold, second best underlined. Ties share the mark.}
\label{tab:seen}
\centering
\resizebox{\columnwidth}{!}{\begin{tabular}{@{}lrrrrrrrrrr@{}}
\toprule
 & \multicolumn{3}{c}{$N{=}2$}
 & \multicolumn{3}{c}{$N{=}3$}
 & \multicolumn{3}{c}{$N{=}4$} \\
\cmidrule(lr){2-4}
\cmidrule(lr){5-7}
\cmidrule(l){8-10}
Method
 & PSNR & SSIM & LPIPS
 & PSNR & SSIM & LPIPS
 & PSNR & SSIM & LPIPS \\
\midrule
Restormer & \underline{24.91} & \underline{0.796} & 0.272 & \underline{22.23} & 0.686 & 0.488 & \underline{20.44} & 0.549 & 0.670 \\
Ada4DIR & 21.03 & 0.760 & 0.312 & 20.00 & 0.662 & 0.492 & 19.06 & 0.551 & 0.679 \\
PhyDAE & 19.03 & 0.746 & \textbf{0.254} & 19.93 & \underline{0.687} & \textbf{0.419} & 19.10 & \textbf{0.593} & \textbf{0.592} \\
CoRE-UIR & 20.55 & 0.748 & 0.267 & 20.04 & 0.673 & \underline{0.438} & 19.45 & \underline{0.578} & \underline{0.593} \\
Ours & \textbf{26.27} & \textbf{0.834} & \underline{0.255} & \textbf{22.77} & \textbf{0.698} & 0.442 & \textbf{21.07} & 0.572 & 0.626 \\
\bottomrule\end{tabular}}
\end{table}

The AIO baselines were not trained on cloud or low resolution, so a natural question is whether EORestore-Agent gains its lead from those unseen types alone. Table~\ref{tab:seen} restricts the comparison to combinations composed entirely of the four degradation types covered by AIO training (haze, noise, blur, low light), covering $N{=}2$ to $4$. It includes the three remote-sensing AIO models and Restormer, the strongest baseline at $N \geq 2$ in Table~\ref{tab:main}. EORestore-Agent leads PSNR at all three values of $N$, with Restormer second in each case: 26.27 dB vs.\ 24.91 dB at $N{=}2$, 22.77 vs.\ 22.23 at $N{=}3$, and 21.07 vs.\ 20.44 at $N{=}4$. The margins of 1.36, 0.54, and 0.63 dB show that the lead holds on combinations where the baselines face no unseen degradation types. They are smaller than the aggregate margins in Table~\ref{tab:main}, which also reflect cloud and low resolution. No unseen type is involved here, so the remaining lead is consistent with a contribution from iterative perception, the merged candidate pool, and scored selection.

\subsection{Ablation Study}
\label{sec:exp_ablation}

\begin{table*}[t]
\caption{Ablation on the Full Benchmark (270 Tiles). PSNR (dB, $\uparrow$), SSIM ($\uparrow$), and LPIPS ($\downarrow$) are averaged over all combinations at each $N$. Selection signal: $\hat{\Delta}_{P,S,L}$ gates on all three predicted changes, $\hat{\Delta}_{P}$ and $\hat{\Delta}_{P,S}$ drop scorer heads, and X-diff ranks candidates by the change in the no-reference score X. Fallback $\times$ removes the input image from the candidate set. Best in bold, second best underlined; ties share the mark.}
\label{tab:ablation}
\centering
\footnotesize
\setlength{\tabcolsep}{4.5pt}
\renewcommand{\arraystretch}{1.1}
\begin{tabular*}{\textwidth}{@{\extracolsep{\fill}}llcrrrrrrrrr@{}}
\toprule
 & & & \multicolumn{3}{c}{$N{=}1$} & \multicolumn{3}{c}{$N{=}2$} & \multicolumn{3}{c}{$N{=}3$} \\
\cmidrule(lr){4-6} \cmidrule(lr){7-9} \cmidrule(l){10-12}
Perception & Selection signal & Fallback & PSNR & SSIM & LPIPS & PSNR & SSIM & LPIPS & PSNR & SSIM & LPIPS \\
\midrule
Single-label & $\hat{\Delta}_{P,S,L}$ & $\checkmark$ & 33.58 & \underline{0.881} & \underline{0.184} & 25.90 & 0.769 & \underline{0.365} & 21.89 & 0.646 & \underline{0.546} \\
Multi-label & $\hat{\Delta}_{P}$ & $\checkmark$ & \underline{33.66} & 0.874 & 0.194 & \textbf{26.05} & 0.762 & 0.376 & \underline{22.06} & 0.640 & 0.556 \\
Multi-label & $\hat{\Delta}_{P,S}$ & $\checkmark$ & \textbf{33.67} & 0.880 & 0.194 & \textbf{26.05} & \underline{0.770} & 0.370 & \textbf{22.07} & \underline{0.647} & 0.551 \\
Multi-label & $\hat{\Delta}_{P}$ & $\times$ & 33.50 & 0.864 & 0.210 & \underline{25.96} & 0.754 & 0.382 & 21.97 & 0.629 & 0.564 \\
\midrule
Multi-label & NIQE-diff & $\checkmark$ & 29.42 & 0.844 & 0.215 & 23.42 & 0.730 & 0.369 & 18.94 & 0.596 & 0.556 \\
Multi-label & MUSIQ-diff & $\checkmark$ & 29.10 & 0.802 & 0.247 & 22.40 & 0.671 & 0.416 & 18.83 & 0.552 & 0.587 \\
Multi-label & TOPIQ-diff & $\checkmark$ & 29.82 & 0.788 & 0.255 & 24.10 & 0.688 & 0.405 & 20.38 & 0.567 & 0.574 \\
\midrule
Multi-label & $\hat{\Delta}_{P,S,L}$ (full) & $\checkmark$ & \underline{33.66} & \textbf{0.882} & \textbf{0.182} & \textbf{26.05} & \textbf{0.771} & \textbf{0.360} & \textbf{22.07} & \textbf{0.650} & \textbf{0.540} \\
\midrule[\heavyrulewidth]
 & & & \multicolumn{3}{c}{$N{=}4$} & \multicolumn{3}{c}{$N{=}5$} & \multicolumn{3}{c}{$N{=}6$} \\
\cmidrule(lr){4-6} \cmidrule(lr){7-9} \cmidrule(l){10-12}
Perception & Selection signal & Fallback & PSNR & SSIM & LPIPS & PSNR & SSIM & LPIPS & PSNR & SSIM & LPIPS \\
\midrule
Single-label & $\hat{\Delta}_{P,S,L}$ & $\checkmark$ & 20.19 & 0.544 & \underline{0.658} & 18.87 & 0.502 & 0.732 & 18.47 & 0.492 & 0.763 \\
Multi-label & $\hat{\Delta}_{P}$ & $\checkmark$ & \textbf{20.58} & 0.538 & 0.672 & \textbf{19.44} & 0.489 & 0.740 & \textbf{19.00} & 0.478 & 0.777 \\
Multi-label & $\hat{\Delta}_{P,S}$ & $\checkmark$ & \underline{20.56} & \underline{0.547} & 0.662 & \textbf{19.44} & \underline{0.506} & 0.734 & \underline{18.98} & \underline{0.497} & 0.765 \\
Multi-label & $\hat{\Delta}_{P}$ & $\times$ & 20.49 & 0.526 & 0.679 & \underline{19.32} & 0.480 & 0.754 & 18.91 & 0.469 & 0.782 \\
\midrule
Multi-label & NIQE-diff & $\checkmark$ & 18.51 & 0.518 & 0.681 & 17.38 & 0.488 & \underline{0.731} & 16.47 & 0.483 & 0.768 \\
Multi-label & MUSIQ-diff & $\checkmark$ & 17.82 & 0.442 & 0.719 & 17.39 & 0.422 & 0.760 & 17.67 & 0.441 & 0.761 \\
Multi-label & TOPIQ-diff & $\checkmark$ & 18.60 & 0.446 & 0.718 & 17.47 & 0.425 & 0.748 & 17.85 & 0.450 & \underline{0.750} \\
\midrule
Multi-label & $\hat{\Delta}_{P,S,L}$ (full) & $\checkmark$ & \underline{20.56} & \textbf{0.550} & \textbf{0.649} & \textbf{19.44} & \textbf{0.509} & \textbf{0.718} & \underline{18.98} & \textbf{0.500} & \textbf{0.745} \\
\bottomrule
\end{tabular*}
\end{table*}

The upper block of Table~\ref{tab:ablation} isolates each component of EORestore-Agent. The first row replaces multi-label perception with single-label, restricting each step's candidate pool to tools of one predicted degradation type and disabling the merged pool. The PSNR gap relative to the full system grows with $N$, from 0.08~dB at $N{=}1$ to 0.57~dB at $N{=}5$ and 0.51~dB at $N{=}6$, with SSIM and LPIPS following the same trend. The merged pool matters most when several degradation types are present and the best candidate may come from a tool outside the dominant type. The next two rows remove scorer heads from the three-metric gate. With only $\hat{\Delta}_P$, PSNR stays within 0.02~dB of the full system and is marginally higher at $N{=}4$ and $N{=}6$ (20.58 vs.\ 20.56 and 19.00 vs.\ 18.98). The gate can only reject candidates, not add them, so filtering by SSIM and LPIPS occasionally removes the PSNR-best option. Yet SSIM and LPIPS worsen without their respective heads. At $N{=}6$, the $\hat{\Delta}_P$-only gate drops SSIM from 0.500 to 0.478 and raises LPIPS from 0.745 to 0.777. Adding $\hat{\Delta}_S$ recovers SSIM to within 0.003, and the full three-head gate recovers LPIPS. Each head guards its own quality dimension, so no accepted action is predicted to trade improvement in one metric for regression in another. The fourth row removes the input fallback. The input image is no longer a candidate, so the agent cannot abstain and commits the candidate with the largest predicted PSNR gain at every step. This row keeps only the $\hat{\Delta}_P$ head, so it differs from the second row only in the ability to abstain. Compared to the second row, PSNR drops by 0.09 to 0.16~dB, SSIM by 0.008 to 0.012, and LPIPS worsens by 0.005 to 0.016 across all six values of $N$. The decline is small but consistent: removing the fallback lowers PSNR on 33 of the 34 combinations. The small size agrees with the step-level diagnosis in Section~\ref{sec:exp_trajectory}, where harmful steps concentrate near the acceptance threshold and change PSNR only slightly. Comparing the two blocks clarifies the division of roles: removing the $\hat{\Delta}_S$ and $\hat{\Delta}_L$ heads leaves PSNR untouched, while replacing the learned scorer with no-reference (NR) difference signals (lower block) costs 1.1 to 4.6~dB. The PSNR gain comes from the learned $\hat{\Delta}_P$ prediction, and the three-metric gate adds SSIM and LPIPS protection at negligible PSNR cost.

The lower block of Table~\ref{tab:ablation} tests whether no-reference quality metrics can replace the learned relative quality scorer as the step-level gating signal. Each row substitutes only the decision signal while keeping the perception layer and toolbox fixed. Three NR-difference baselines are evaluated, built on NIQE~\cite{niqe2013} (non-parametric), MUSIQ~\cite{musiq2021} (learned), and TOPIQ~\cite{chen2024topiq} (learned), and denoted NIQE-diff, MUSIQ-diff, and TOPIQ-diff. Our scorer leads PSNR at every $N$, by 3.84 dB over the best NR baseline (TOPIQ-diff) at $N{=}1$ and by 1.13 dB at $N{=}6$. NIQE-diff comes closest in SSIM, within 0.017 at $N{=}6$, but falls behind in PSNR by 4.24 dB at $N{=}1$. The three NR-difference signals favor different metrics. TOPIQ-diff gives the best PSNR among them at every $N$, NIQE-diff gives the best SSIM, and neither matches our scorer in PSNR. Predicting the quality change from an input-candidate pair is more reliable for per-step gating than differencing absolute NR scores, which neither indicate whether a step has improved the image~\cite{hu2022hierarchical} nor support fine-grained comparison among candidates~\cite{sheng2026fine}.

\subsection{Real-World Composite Degradation}
\label{sec:exp_real}

\begin{figure*}[t]
\centering
\includegraphics[width=\textwidth]{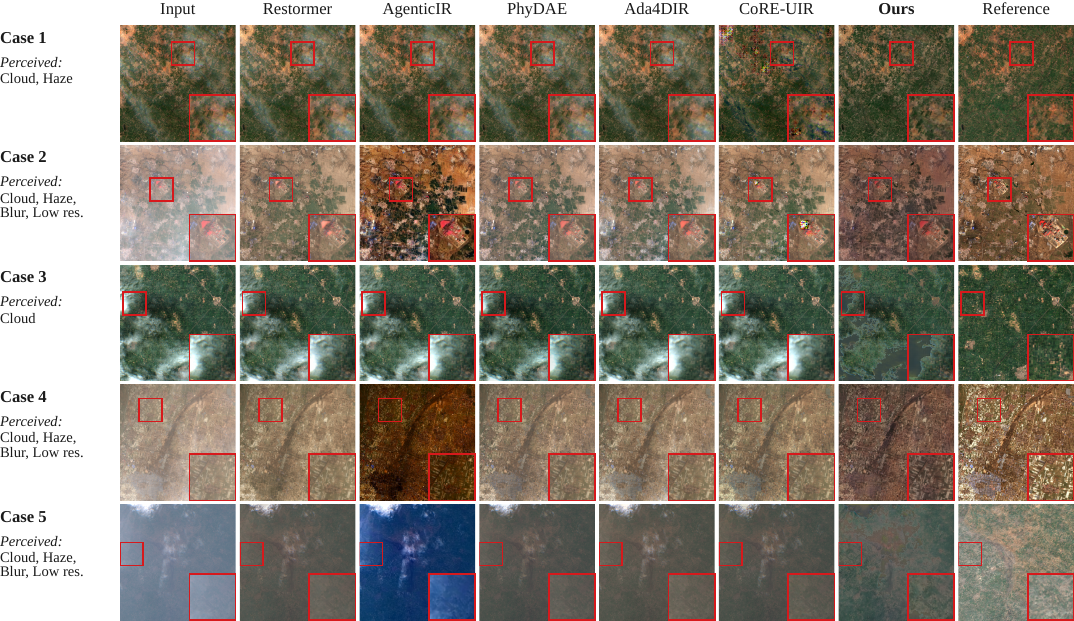}
\caption{Real-world cases on Sentinel-2 L2A tiles. The reference is the L2A acquisition of the same area from a nearby cloud-free date, for visual comparison only (no full-reference metric). Row labels list the degradation set perceived at step 1. The red box marks the region enlarged at the bottom right of each cell.}
\label{fig:real}
\end{figure*}

Fig.~\ref{fig:real} presents five cases of naturally occurring composite degradation on Sentinel-2 L2A tiles (ESA true-color images of 512 $\times$ 512 pixels at an effective 20 m ground sampling, about 10 km per side). EORestore-Agent uses the weights trained on synthetic Landsat-8 data without retraining. The inputs carry no degradation labels. The reference column is the L2A acquisition of the same area from a nearby cloud-free date, 10 to 30 days from the input, for visual comparison only. Both the input and reference are L2A surface reflectance products, but acquisition dates and atmospheric conditions differ, so no full-reference metric is computed.

In case 1, EORestore-Agent removes thin cirrus streaks from a vegetated plain and matches the reference tone. In case 2, haze covers roughly half the frame over mixed urban and agricultural land. EORestore-Agent clears it with spatially varying correction, keeping field boundaries and rooftop detail. In case 3, EORestore-Agent removes discrete cumulus clouds, but cloud outlines remain and the filled regions cannot be verified against the reference. In case 4, EORestore-Agent strips dense haze over an urban area and recovers street and rail geometry, though the output is softer than the reference. Case 5 has the heaviest haze in the set. EORestore-Agent produces the largest visible change, but the recovered content is patchy.

PhyDAE and Ada4DIR produce outputs nearly identical to the input on every case. CoRE-UIR is the most active remote-sensing baseline and the one that fails visibly, with color blocks on case 1 and speckle on case 2. Restormer dehazes competently on cases 2 and 4 but leaves cloud untouched. AgenticIR applies strong color casts, orange on cases 2 and 4 and blue on case 5. EORestore-Agent is the only method in the set that removes an opaque cloud.

The perceived set at step 1 reads haze and cloud on case 1, cloud alone on case 3, and low resolution, haze, cloud, and blur on the other three. The set is stable across hazy inputs. Low resolution is reported on three of the five cases, consistent with the reduced 20 m sampling of these tiles.

The degradation in these tiles is real atmosphere that survived ESA's L2A atmospheric correction. EORestore-Agent removes the residual veil but keeps more blue than the reference on some cases and shifts toward red on others. The scorer, trained on synthetic degradation, over-rewards dehazing on real imagery. Restormer returns brighter panels on cases 2 and 4. The relevant comparison is what each method removes from the degraded input, not which output appears brightest.

\subsection{Step-Level Trajectory Diagnosis}
\label{sec:exp_trajectory}

\begin{figure*}[t]
\centering
\includegraphics[width=\textwidth]{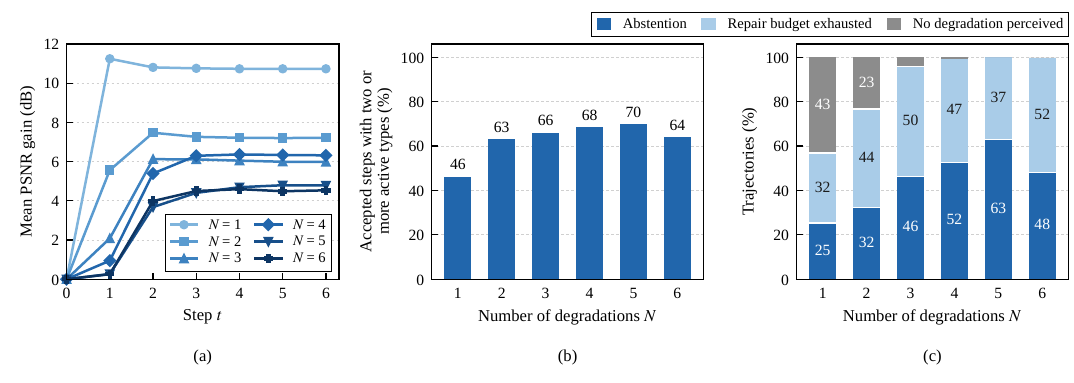}
\caption{Trajectory statistics. (a) Mean true PSNR gain at each step, one curve per $N$. Terminated trajectories hold their final value. (b) Share of accepted steps with two or more active types. (c) Share of trajectories that end by abstention, by exhausting the repair budget, or because no degradation is perceived.}
\label{fig:trajectory}
\end{figure*}

\begin{figure*}[t]
\centering
\includegraphics[width=\textwidth]{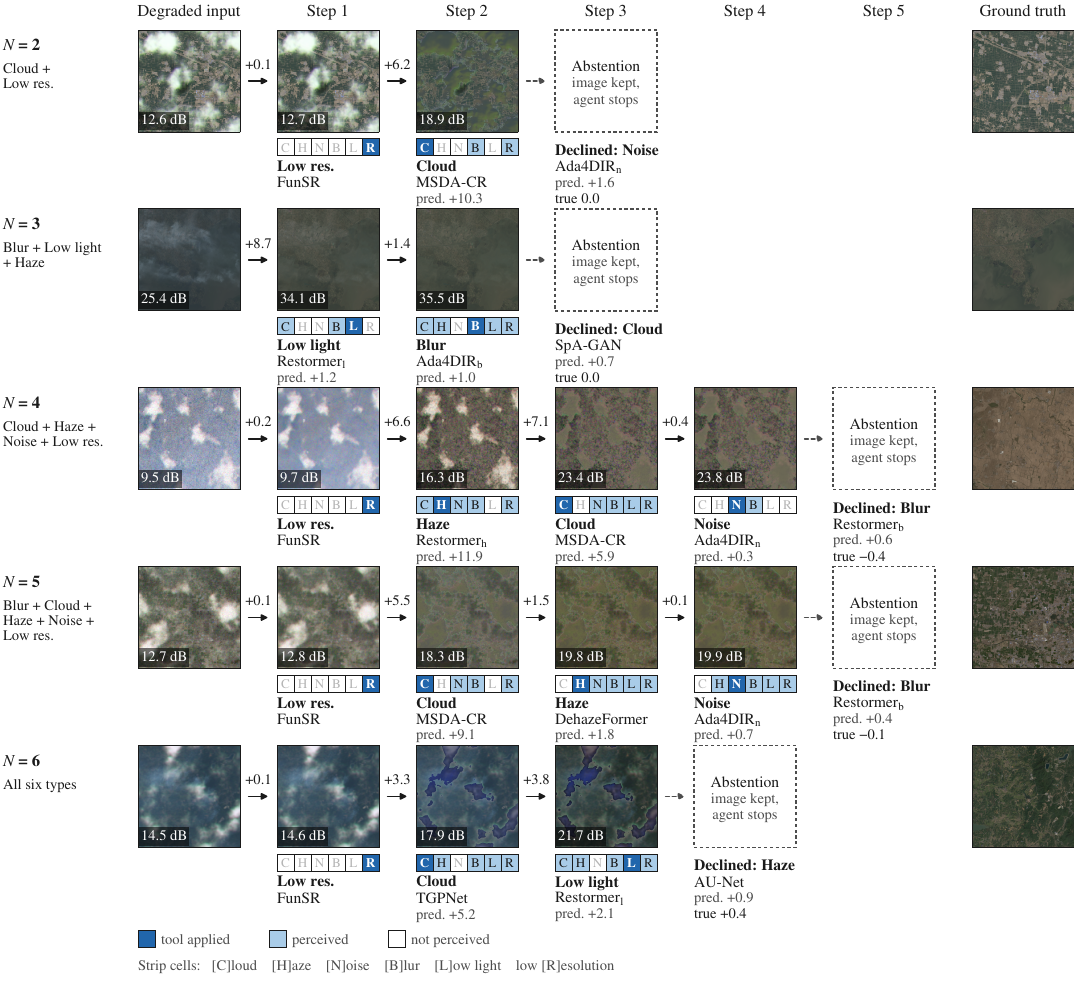}
\caption{Restoration trajectories, one per $N$ from 2 to 6. Rows 2 to 4 use the tiles of Fig.~\ref{fig:qualitative}, rows 1 and 5 a different tile of the same combination. Each solid arrow is an accepted step, labeled with its true PSNR change in dB. Under each frame, the strip marks the perceived types and the type of the applied tool, followed by that type, the tool, and its predicted PSNR change. Each trajectory ends with an abstention (dashed box). No tool is predicted to raise PSNR without worsening SSIM or LPIPS, so the agent keeps the current image and stops. Below the box is the declined tool, the one with the highest predicted PSNR gain. The three-metric gate rejected it because its predicted SSIM or LPIPS change was unfavorable.}
\label{fig:filmstrip}
\end{figure*}

Fig.~\ref{fig:trajectory} reports statistics over the full-system trajectories behind the Ours row of Table~\ref{tab:main}. This step-level diagnosis asks three questions. Do decisions made on predicted changes add up to a true gain? How much does each part of the selection rule of Section~\ref{sec:method_scoring} contribute? Is the merged pool of Section~\ref{sec:method_pool} used during restoration?

Panel (a) tests the premise of Section~\ref{sec:method_overview}, that decisions made on predicted changes add up to a true fidelity gain. The agent never observes true PSNR, yet the mean true PSNR gain rises at every $N$ and reaches more than 90 percent of its peak by the third step. The curves then flatten because the agent stops on its own. The curves are not strictly monotone. After the peak, the mean gain falls by 0.51~dB at $N{=}1$ and by 0.01 to 0.26~dB at larger $N$, which is at most 5 percent of the peak gain. Wrongly accepted steps therefore cost little compared with what the accepted steps deliver, and errors do not compound along the trajectory. The final gain, Ours minus the degraded input in Table~\ref{tab:main}, falls from 10.7~dB at $N{=}1$ to 4.5~dB at $N{=}6$.

\begin{table}[t]
\caption{Single-step comparison of selection rules on the same candidate pools. Harmful rate: share of steps whose chosen candidate lowers true PSNR. Gain: mean true PSNR change of the chosen candidate, over the steps where the full rule accepts a tool. Oracle picks the candidate with the highest true gain. The last column is the gain of Ours as a share of the oracle gain.}
\label{tab:selector}
\centering
\resizebox{\columnwidth}{!}{\begin{tabular}{@{}crrrrrrrr@{}}
\toprule
 & \multicolumn{3}{c}{Harmful rate (\%) $\downarrow$} & \multicolumn{4}{c}{Gain (dB) $\uparrow$} & \\
\cmidrule(lr){2-4} \cmidrule(lr){5-8}
$N$ & Random & Ranking & Ours & Random & Ranking & Ours & Oracle & \% of oracle \\
\midrule
1 & 46.7 & 40.0 & 27.9 & 3.97 & 7.68 & 7.69 & 8.28 & 93 \\
2 & 43.2 & 30.4 & 24.7 & 1.04 & 3.63 & 3.57 & 4.40 & 81 \\
3 & 48.9 & 35.5 & 27.1 & 0.07 & 1.87 & 1.84 & 2.57 & 72 \\
4 & 48.2 & 33.8 & 26.6 & 0.29 & 1.65 & 1.72 & 2.32 & 74 \\
5 & 41.4 & 28.2 & 21.4 & 0.29 & 1.30 & 1.22 & 2.16 & 56 \\
6 & 42.4 & 29.4 & 24.3 & 0.05 & 1.00 & 0.96 & 1.67 & 57 \\
\bottomrule\end{tabular}}
\end{table}

Table~\ref{tab:selector} separates the contributions of the two parts of the selection rule. Three rules are applied to the same candidate pools: a uniform random pick, the predicted-PSNR ranking without the gate, and the full rule, which adds the three-metric gate and abstention. An oracle that picks the candidate with the highest true gain bounds any single-step rule. We call a step harmful when the chosen candidate lowers true PSNR. An abstention leaves the image unchanged and counts as not harmful. A random pick is harmful on almost half of the steps and gains almost nothing at $N \geq 3$, so the merged pool alone does not help. The ranking brings the gain. With the ranking alone, the mean true gain reaches 60 to 93 percent of the oracle, and the harmful rate drops by 7 to 14 points. This share broadly falls as $N$ grows. The gain columns cover only the steps where the full rule accepts a tool. On these steps the ranking-only and full rules differ only where the SSIM and LPIPS constraints reject the top-ranked candidate, which happens about 10 percent of the time. The replacement targets SSIM and LPIPS, and its net effect on PSNR is close to zero: the two gains differ by at most 0.08~dB. Abstention acts on the remaining steps, where it removes 6 to 12 points of harmful rate. The replacement changes the harmful rate by at most 2 points in either direction. Its benefit shows at the trajectory level: removing it lowers all three metrics in the end-to-end ablation (Table~\ref{tab:ablation}). It also declines the right steps: on more than half of the declined steps at every $N$, the top-ranked candidate would have been harmful.

The full rule still accepts harmful steps. Counted over accepted steps only, 25 to 33 percent are harmful, against 21 to 28 percent over all steps in Table~\ref{tab:selector}. This residual is a calibration problem at the acceptance threshold, not a ranking problem. The scorer overestimates gains, most visibly for large predictions in Fig.~\ref{fig:filmstrip}: $+10.3$, $+11.9$, and $+9.1$~dB predicted against $+6.2$, $+6.6$, and $+5.5$~dB true. On accepted steps, the mean predicted gain exceeds the mean true gain by 0.5~dB at $N{=}1$ and by 2.1~dB at $N{=}6$, so candidates with a small true loss can clear the zero threshold. Accordingly, harmful steps concentrate near the threshold. Among accepted steps with a predicted gain above 3~dB, 7 to 21 percent are harmful. Among those below 1~dB, about half are. Deblurring tools are the largest single source of harmful acceptances. The harmful steps are small in magnitude and the positive steps outweigh them, which is why the curves in panel (a) lose so little after their peak.

Panel (b) checks whether the merged pool is exercised in practice. At $N \geq 2$, 63 to 70 percent of accepted steps are selected from a pool that merges at least two active degradation types, so most decisions are taken across types and not within one. The rate is 46 percent at $N{=}1$, where only one type is present, because the perception layer reports 1.79 types on average at the first step. This exposes the cost of the merged pool: it passes perception false positives on to the scorer. At $N{=}1$, 37 percent of accepted steps apply a tool for a type absent from the image, and 84 percent of these are harmful, against 4 percent when the tool matches the true type. This is consistent with the post-peak drop in panel (a) being largest at $N{=}1$. The share of such steps falls as $N$ grows, because fewer types are absent.

A trajectory can end in three ways (Algorithm~\ref{alg:pipeline}): the perception layer reports no degradation, every perceived type has been closed by abstention, or every perceived type has used its repair budget. Panel (c) shows that every trajectory is ended by one of these mechanisms and that the dominant one shifts with $N$. At $N{=}1$, 43 percent of trajectories end because the perception layer reports no remaining degradation. This route nearly vanishes by $N{=}4$. Abstention takes over, rising from 25 percent of trajectories at $N{=}1$ to 63 percent at $N{=}5$ and 48 percent at $N{=}6$, and repair-budget exhaustion ends the rest. At high $N$ the agent therefore stops mainly because the scorer finds no tool worth applying, not because the image is judged clean.

Fig.~\ref{fig:filmstrip} shows how these mechanisms act on five individual trajectories, one per $N$ from 2 to 6. Row 2 illustrates the benefit of the merged pool. At the first step the perception layer reports cloud, blur, and low light, where cloud is a false positive and haze is missed. A pipeline tied to a single perceived type could have searched only cloud tools. The merged pool holds the tools of all three types, and the scorer picks a low-light tool that is also the best of the eight candidates in true gain, 8.7~dB in one step. Rows 3 and 4 show that re-perceiving and re-ranking at each step fixes the restoration order implicitly. The two images share four degradation types, yet the agent dehazes before removing cloud on row 3 and does the opposite on row 4, because the ranking is recomputed from the input image at every step. Both trajectories gain more than 7~dB. Every row ends with an abstention, shown as a dashed box with the tool it declines. The last perception round reports types that are absent from the image or that no tool improves. No tool is predicted to raise PSNR without worsening SSIM or LPIPS, and the agent stops. Abstaining forgoes almost no gain. The declined tools would have changed true PSNR by $-0.4$ to $+0.4$~dB. On row 1, a rule based on predicted PSNR alone would have accepted a denoising step with a predicted gain of 1.6~dB and a true gain of zero.

\section{Conclusion}
\label{sec:conclusion}

We presented EORestore-Agent, a restoration agent for optical remote sensing images with composite degradations. By merging the tool pools of all perceived degradation types into a merged candidate pool and scoring candidates by their predicted quality change across PSNR, SSIM, and LPIPS, the framework selects one restoration action per step and stops when no tool is predicted to improve the image. On a synthetic benchmark spanning composites of two to six degradation types, this design improves PSNR by 2.3 to 3.2~dB over the strongest baseline, whereas all-in-one models converge to a narrow performance band regardless of architecture and natural-image restoration agents produce outputs below the degraded input in fidelity. Ablation studies trace the PSNR advantage to the learned relative quality scorer, with the merged pool and additional metric gates contributing complementary gains in structural and perceptual quality. Trajectory analysis shows that most improvement is delivered within the first few steps, and that the remaining harmful decisions stem primarily from scorer overestimation and perception errors. Qualitative results on Sentinel-2 imagery confirm that the agent transfers to real atmospheric degradation without retraining. Future work will focus on calibrating the acceptance threshold under real degradation distributions and extending evaluation to multispectral imagery and downstream remote sensing tasks.




\clearpage
{\appendices

\section{Perception Prompt Templates}
\label{app:prompts}

\subsection{Prompt Template}

\begin{tcolorbox}[
  enhanced, frame hidden, sharp corners,
  colback=black!3, borderline west={1pt}{0pt}{black!35},
  boxrule=0pt, left=8pt, right=8pt, top=7pt, bottom=7pt]
You are an expert remote sensing image quality analyst. Examine the image.

\medskip
Question: Is the degradation ``\{type\}'' present in this image?

\medskip
Definition of ``\{type\}'': \{definition\}

\medskip
A) Yes, it is clearly present

B) No, it is not present

\medskip
Answer with the single letter A or B, and your reasoning for making the decision.
\end{tcolorbox}

\subsection{Degradation Definitions}

\noindent\textbf{Haze:}
broad atmospheric veil: the image is affected by a smooth, widespread atmospheric veil that reduces contrast and visibility. Colors may look washed out, desaturated, pale, or slightly bluish/gray, while the ground scene remains partly visible. The dominant cue is a broad low-contrast veil, not distinct cloud bodies, cloud patches, cloud wisps, or cloud shadows.

\medskip
\noindent\textbf{Low resolution:}
low spatial resolution / insufficient spatial sampling: the scene appears under-resolved, with too little spatial detail for the expected remote sensing observation. Fine structures such as small buildings, narrow roads, field boundaries, object edges, and high-frequency textures are not clearly resolvable; they may appear missing, merged, jagged, blocky, or overly simplified. The dominant cue is insufficient spatial sampling or pixel-level coarseness, not smooth image blur, haze, noise, or darkness.

\noindent\textbf{Cloud:}
visible cloud contamination: the image contains distinct white or gray cloud bodies, wisps, patches, streaks, or opaque/translucent regions with spatial structure. Clouds may partially or fully obscure the ground and may be accompanied by darker cloud shadows. Structured thin cirrus, cloud wisps, cloud streaks, or cloud patches should be classified as cloud, not haze. Do not classify bright terrain surfaces, rooftops, snow, sand, salt flats, sunglint, or other terrain-aligned bright objects as cloud if they do not visibly obscure the ground.

\medskip
\noindent\textbf{Blur:}
loss of sharpness: the image has reduced sharpness because edges, boundaries, and textures appear smoothly softened, smeared, or fuzzy. Objects are still present at the expected spatial scale, but their contours and fine textures are less crisp. The dominant cue is softened image detail and fuzzy boundaries, not insufficient spatial sampling, haze, noise, or darkness.

\medskip
\noindent\textbf{Noise:}
irregular sensor or acquisition noise: the image contains unnatural fine-scale fluctuations, grain, speckle, or color/intensity variations that are not aligned with real terrain, vegetation, buildings, roads, clouds, or other scene structures. The dominant cue is irregular visual contamination over the image, not natural land-cover texture, urban detail, compression blocks, tiling seams, or resampling artifacts.

\medskip
\noindent\textbf{Low light:}
low light / underexposed appearance: for normal daytime optical remote sensing imagery, the image is globally too dark, with low brightness across large areas, compressed shadow regions, and reduced visibility due to insufficient illumination or poor exposure. Local terrain shadows or cloud shadows alone do NOT count as global dark degradation.

\section{Per-Combination Results}
\label{app:percombo}

Tables~\ref{tab:percombo_n1} to \ref{tab:percombo_n6} report results for each degradation combination. Column labels abbreviate the degradation types: CL cloud, HZ haze, NS noise, BL blur, LL low light, and LR low resolution. A check in the Seen row marks combinations composed entirely of degradation types covered by AIO training. Best and second best among restoration methods are in bold and underlined. The degraded input is not ranked. Ties share the mark.

Table~\ref{tab:percombo_n1} locates the single-degradation gap discussed in Section~\ref{sec:exp_main}. On the four types covered by AIO training, the best AIO model exceeds EORestore-Agent by 1.06 to 6.87~dB. On the two types outside that training, the result reverses. EORestore-Agent leads by 4.88~dB on cloud and by 0.86~dB on low resolution, where it is the only method above the degraded input (31.88 vs.\ 31.23~dB). The deficit on the seen types is consistent with the perception false positives at $N{=}1$ (Section~\ref{sec:exp_trajectory}).

From $N{=}2$ on, EORestore-Agent has the highest PSNR on all 28 combinations Tables~\ref{tab:percombo_n2} to \ref{tab:percombo_n6}). On the six seen combinations at $N{=}2$ to 4, the lead over the best baseline is 0.13 to 2.37~dB, in line with Table~\ref{tab:seen}. The three combinations that pair cloud with low resolution at $N{=}2$ and 3 give the largest leads, 5.72 to 6.79~dB. At $N{=}5$ and 6, every combination contains cloud or low resolution, and the lead is 0.79 to 3.50~dB.

Among the AIO models, Restormer is the best on 18 of the 28 combinations. AirNet, the weakest AIO at $N{=}1$ (28.13~dB), is the best on seven of the remaining ten, all of which contain cloud. The single-degradation ranking therefore does not carry over to composites, consistent with the shared performance ceiling discussed in Section~\ref{sec:exp_main}.

Blur with low resolution at $N{=}2$ is the only combination on which every method, including EORestore-Agent, falls below the degraded input (32.01~dB). EORestore-Agent loses the least, 0.25~dB. This agrees with Section~\ref{sec:exp_trajectory}, where the remaining harmful steps are small and concentrate near the acceptance threshold.

Table~\ref{tab:percombo_n6} applies all six types in four orders. EORestore-Agent reaches 17.96 to 19.68~dB, and its lead over the best baseline stays between 2.11 and 2.91~dB, so the advantage holds across all four degradation orders. Order~4, which applies low resolution first, gives the lowest PSNR for both EORestore-Agent and the best baseline.

\begin{table*}[p]
\caption{Per-combination results, $N{=}1$. Column labels: Cl cloud, Hz haze, Ns noise, Bl blur, LL low light, LR low resolution. A check in the Seen row marks combinations composed entirely of degradation types covered by AIO training. Method references are given in Table~\ref{tab:main}.}
\label{tab:percombo_n1}
\centering
\footnotesize
\renewcommand{\arraystretch}{0.95}
\resizebox{\textwidth}{!}{\begin{tabular}{@{}lrrrrrrrrrrrrrrrrrr@{}}
\toprule
 & \multicolumn{3}{c}{Bl} & \multicolumn{3}{c}{LL} & \multicolumn{3}{c}{Hz} & \multicolumn{3}{c}{Ns} & \multicolumn{3}{c}{Cl} & \multicolumn{3}{c}{LR} \\
\cmidrule(lr){2-4} \cmidrule(lr){5-7} \cmidrule(lr){8-10} \cmidrule(lr){11-13} \cmidrule(lr){14-16} \cmidrule(l){17-19}
Method & PSNR & SSIM & LPIPS & PSNR & SSIM & LPIPS & PSNR & SSIM & LPIPS & PSNR & SSIM & LPIPS & PSNR & SSIM & LPIPS & PSNR & SSIM & LPIPS \\
\midrule
Seen & \multicolumn{3}{c}{$\checkmark$} & \multicolumn{3}{c}{$\checkmark$} & \multicolumn{3}{c}{$\checkmark$} & \multicolumn{3}{c}{$\checkmark$} & \multicolumn{3}{c}{} & \multicolumn{3}{c}{} \\
\midrule
Degraded input & 34.10 & 0.812 & 0.402 & 15.79 & 0.590 & 0.233 & 14.18 & 0.799 & 0.249 & 25.19 & 0.458 & 0.565 & 17.11 & 0.816 & 0.318 & 31.23 & 0.759 & 0.496 \\
\midrule
\multicolumn{19}{@{}l}{\emph{Natural-image all-in-one}} \\
PromptIR & 36.43 & 0.904 & 0.171 & 40.93 & 0.995 & 0.005 & 36.42 & 0.982 & 0.025 & 35.91 & 0.895 & 0.154 & 17.24 & 0.815 & 0.305 & 23.97 & 0.734 & 0.499 \\
Restormer & 36.24 & 0.892 & 0.211 & 41.91 & 0.995 & 0.004 & 37.07 & 0.983 & 0.024 & 35.74 & 0.895 & 0.138 & 17.10 & 0.819 & 0.305 & 27.78 & 0.748 & 0.492 \\
NAFNet & 36.61 & 0.904 & 0.181 & \underline{47.15} & \textbf{0.997} & \textbf{0.002} & \underline{40.20} & \underline{0.987} & \underline{0.015} & 36.03 & 0.898 & 0.133 & 18.10 & \underline{0.823} & 0.298 & 27.53 & 0.748 & 0.499 \\
DGUNet & 36.34 & 0.901 & 0.154 & 36.86 & 0.991 & 0.010 & 36.48 & 0.982 & 0.026 & 35.81 & 0.893 & 0.153 & 16.34 & 0.801 & 0.316 & 28.56 & 0.744 & 0.496 \\
IDR & 35.70 & 0.886 & 0.202 & 34.54 & 0.986 & 0.014 & 32.68 & 0.972 & 0.046 & 35.53 & 0.891 & 0.132 & 17.16 & 0.803 & 0.312 & 25.32 & 0.731 & 0.511 \\
SRResNet-S+AP & 36.50 & 0.906 & 0.163 & 39.25 & 0.994 & 0.007 & 33.86 & 0.974 & 0.044 & 35.99 & 0.898 & 0.138 & 17.16 & 0.819 & 0.295 & 27.10 & 0.741 & 0.501 \\
Uformer-S+AP & \textbf{37.14} & \textbf{0.915} & 0.162 & 40.98 & \underline{0.996} & 0.004 & 38.02 & 0.985 & 0.020 & \underline{36.09} & \underline{0.899} & 0.136 & 17.45 & \textbf{0.824} & \underline{0.290} & \underline{31.02} & 0.750 & \underline{0.487} \\
Restormer-S+AP & 37.02 & 0.907 & 0.166 & 43.48 & \underline{0.996} & \underline{0.003} & 38.61 & 0.986 & 0.019 & 35.57 & 0.886 & \underline{0.103} & 17.30 & \underline{0.823} & 0.301 & 28.52 & \underline{0.752} & \underline{0.487} \\
AirNet & 32.47 & 0.834 & 0.324 & 27.20 & 0.965 & 0.035 & 27.71 & 0.943 & 0.082 & 35.19 & 0.886 & 0.153 & 18.22 & 0.814 & 0.321 & 27.98 & 0.745 & 0.509 \\
\midrule
\multicolumn{19}{@{}l}{\emph{Remote-sensing all-in-one}} \\
PhyDAE & 35.25 & 0.869 & \underline{0.098} & 41.82 & \underline{0.996} & \underline{0.003} & 36.74 & 0.985 & 0.020 & 35.46 & 0.894 & 0.105 & 17.65 & 0.822 & 0.300 & 26.73 & 0.741 & 0.501 \\
Ada4DIR & \underline{37.08} & \underline{0.914} & 0.151 & 43.63 & \underline{0.996} & \underline{0.003} & 39.34 & \underline{0.987} & 0.016 & \textbf{36.13} & \underline{0.899} & 0.145 & 17.10 & 0.817 & 0.307 & 27.93 & 0.745 & 0.496 \\
CoRE-UIR & 36.99 & \underline{0.914} & \textbf{0.049} & \textbf{47.42} & \textbf{0.997} & \textbf{0.002} & \textbf{41.08} & \textbf{0.989} & \textbf{0.011} & 36.03 & \textbf{0.901} & \textbf{0.069} & \underline{18.48} & 0.789 & 0.307 & 27.65 & 0.733 & 0.522 \\
\midrule
\multicolumn{19}{@{}l}{\emph{Natural-image systems (zero-shot)}} \\
AutoDIR & 23.46 & 0.606 & 0.383 & 15.70 & 0.732 & 0.210 & 14.20 & 0.733 & 0.322 & 23.74 & 0.728 & 0.245 & 15.89 & 0.599 & 0.393 & 18.78 & 0.555 & 0.691 \\
AgenticIR & 17.45 & 0.595 & 0.435 & 19.15 & 0.688 & 0.259 & 14.67 & 0.553 & 0.371 & 19.08 & 0.579 & 0.495 & 13.86 & 0.583 & 0.448 & 16.83 & 0.340 & 0.501 \\
4KAgent & 13.00 & 0.316 & 0.589 & 16.18 & 0.529 & 0.387 & 14.93 & 0.510 & 0.428 & 16.39 & 0.511 & 0.464 & 12.16 & 0.357 & 0.584 & 14.14 & 0.326 & 0.566 \\
JarvisIR & 15.44 & 0.557 & 0.525 & 16.72 & 0.550 & 0.434 & 13.53 & 0.563 & 0.469 & 15.77 & 0.507 & 0.412 & 13.24 & 0.511 & 0.515 & 16.15 & 0.484 & 0.561 \\
\midrule
\multicolumn{19}{@{}l}{\emph{EORestore-Agent}} \\
Ours & 35.64 & 0.888 & 0.157 & 41.82 & 0.984 & 0.011 & 34.21 & 0.954 & 0.059 & 35.07 & 0.890 & 0.133 & \textbf{23.36} & 0.816 & \textbf{0.271} & \textbf{31.88} & \textbf{0.762} & \textbf{0.461} \\
\bottomrule\end{tabular}}
\end{table*}

\begin{table*}[p]
\caption{Per-combination results, $N{=}2$. Column labels: Cl cloud, Hz haze, Ns noise, Bl blur, LL low light, LR low resolution. A check in the Seen row marks combinations composed entirely of degradation types covered by AIO training. Method references are given in Table~\ref{tab:main}.}
\label{tab:percombo_n2}
\centering
\footnotesize
\renewcommand{\arraystretch}{0.95}
\resizebox{\textwidth}{!}{\begin{tabular}{@{}lrrrrrrrrrrrrrrrrrr@{}}
\toprule
 & \multicolumn{3}{c}{Bl+LL} & \multicolumn{3}{c}{LL+Hz} & \multicolumn{3}{c}{Hz+Ns} & \multicolumn{3}{c}{Bl+LR} & \multicolumn{3}{c}{Cl+Ns} & \multicolumn{3}{c}{Cl+LR} \\
\cmidrule(lr){2-4} \cmidrule(lr){5-7} \cmidrule(lr){8-10} \cmidrule(lr){11-13} \cmidrule(lr){14-16} \cmidrule(l){17-19}
Method & PSNR & SSIM & LPIPS & PSNR & SSIM & LPIPS & PSNR & SSIM & LPIPS & PSNR & SSIM & LPIPS & PSNR & SSIM & LPIPS & PSNR & SSIM & LPIPS \\
\midrule
Seen & \multicolumn{3}{c}{$\checkmark$} & \multicolumn{3}{c}{$\checkmark$} & \multicolumn{3}{c}{$\checkmark$} & \multicolumn{3}{c}{} & \multicolumn{3}{c}{} & \multicolumn{3}{c}{} \\
\midrule
Degraded input & 15.94 & 0.487 & 0.535 & 18.45 & 0.768 & 0.386 & 13.96 & 0.360 & 0.745 & 32.01 & 0.749 & 0.517 & 16.04 & 0.385 & 0.695 & 16.63 & 0.624 & 0.640 \\
\midrule
\multicolumn{19}{@{}l}{\emph{Natural-image all-in-one}} \\
PromptIR & 19.70 & 0.687 & 0.296 & 20.59 & 0.792 & 0.212 & 15.38 & 0.696 & 0.403 & 28.22 & 0.747 & 0.516 & 16.95 & 0.719 & 0.401 & 15.56 & 0.613 & 0.640 \\
Restormer & 26.08 & 0.790 & 0.306 & \underline{26.02} & \underline{0.871} & 0.189 & \underline{22.62} & \underline{0.727} & \underline{0.322} & 29.87 & 0.753 & 0.512 & 16.87 & 0.716 & 0.388 & 15.38 & 0.609 & 0.635 \\
NAFNet & 23.43 & 0.763 & 0.322 & 21.54 & 0.812 & 0.199 & 14.84 & 0.693 & 0.377 & 30.72 & \underline{0.754} & 0.510 & 16.94 & \underline{0.724} & 0.380 & 16.13 & 0.616 & 0.641 \\
DGUNet & 19.07 & 0.645 & 0.291 & 21.55 & 0.827 & 0.216 & 15.62 & 0.675 & 0.397 & 30.54 & 0.749 & 0.509 & 16.88 & 0.709 & 0.409 & 15.41 & 0.609 & 0.642 \\
IDR & 24.56 & 0.777 & 0.357 & 19.38 & 0.765 & 0.233 & 18.17 & 0.576 & 0.458 & 27.22 & 0.738 & 0.526 & 16.98 & 0.719 & 0.383 & 15.78 & 0.597 & 0.658 \\
SRResNet-S+AP & 23.12 & 0.756 & 0.335 & 21.22 & 0.797 & 0.262 & 14.81 & 0.665 & 0.387 & 29.41 & 0.750 & 0.513 & 16.95 & 0.718 & 0.382 & 15.78 & 0.610 & 0.637 \\
Uformer-S+AP & 21.29 & 0.771 & 0.289 & 21.62 & 0.825 & 0.191 & 16.68 & 0.695 & 0.354 & \underline{30.82} & 0.749 & 0.510 & 16.94 & \underline{0.724} & 0.373 & 16.43 & 0.612 & 0.634 \\
Restormer-S+AP & 23.64 & 0.784 & 0.308 & 21.13 & 0.803 & 0.180 & 15.60 & 0.706 & 0.323 & 30.75 & 0.753 & \underline{0.508} & 16.88 & 0.717 & 0.371 & 16.26 & \underline{0.622} & 0.629 \\
AirNet & 25.21 & 0.786 & 0.399 & 20.05 & 0.754 & 0.269 & 18.79 & 0.610 & 0.435 & 28.74 & 0.746 & 0.533 & \underline{17.31} & 0.721 & 0.389 & \underline{16.56} & 0.618 & 0.639 \\
\midrule
\multicolumn{19}{@{}l}{\emph{Remote-sensing all-in-one}} \\
PhyDAE & 21.23 & 0.713 & \underline{0.256} & 20.79 & 0.815 & 0.185 & 15.08 & 0.710 & \textbf{0.321} & 29.71 & 0.752 & 0.513 & 16.92 & 0.723 & 0.364 & 15.34 & 0.608 & 0.643 \\
Ada4DIR & \underline{26.45} & \underline{0.803} & 0.327 & 20.96 & 0.813 & 0.188 & 15.69 & 0.664 & 0.420 & 29.49 & 0.752 & 0.513 & 16.91 & 0.721 & 0.391 & 15.77 & 0.617 & 0.639 \\
CoRE-UIR & 24.04 & 0.753 & 0.263 & 21.09 & 0.816 & \underline{0.179} & 16.52 & 0.676 & 0.360 & 30.04 & 0.735 & 0.524 & 16.94 & \textbf{0.725} & \textbf{0.337} & 15.82 & 0.573 & 0.677 \\
\midrule
\multicolumn{19}{@{}l}{\emph{Natural-image systems (zero-shot)}} \\
AutoDIR & 16.65 & 0.528 & 0.438 & 15.28 & 0.639 & 0.481 & 14.22 & 0.594 & 0.462 & 14.30 & 0.448 & 0.824 & 14.34 & 0.482 & 0.483 & 13.44 & 0.519 & 0.738 \\
AgenticIR & 21.21 & 0.702 & 0.380 & 14.81 & 0.538 & 0.468 & 15.13 & 0.530 & 0.610 & 16.56 & 0.324 & 0.514 & 15.30 & 0.564 & 0.574 & 13.76 & 0.325 & \underline{0.608} \\
4KAgent & 14.87 & 0.506 & 0.467 & 12.63 & 0.423 & 0.497 & 13.58 & 0.585 & 0.488 & 15.26 & 0.313 & 0.562 & 15.87 & 0.584 & 0.487 & 13.74 & 0.314 & 0.627 \\
JarvisIR & 16.11 & 0.540 & 0.560 & 12.79 & 0.541 & 0.556 & 12.60 & 0.523 & 0.543 & 15.16 & 0.528 & 0.589 & 13.29 & 0.470 & 0.541 & 13.69 & 0.438 & 0.646 \\
\midrule
\multicolumn{19}{@{}l}{\emph{EORestore-Agent}} \\
Ours & \textbf{28.82} & \textbf{0.874} & \textbf{0.222} & \textbf{26.21} & \textbf{0.887} & \textbf{0.175} & \textbf{23.79} & \textbf{0.741} & 0.367 & \textbf{31.76} & \textbf{0.756} & \textbf{0.490} & \textbf{22.64} & \underline{0.724} & \underline{0.347} & \textbf{23.08} & \textbf{0.644} & \textbf{0.559} \\
\bottomrule\end{tabular}}
\end{table*}

\begin{table*}[p]
\caption{Per-combination results, $N{=}3$. Column labels: Cl cloud, Hz haze, Ns noise, Bl blur, LL low light, LR low resolution. A check in the Seen row marks combinations composed entirely of degradation types covered by AIO training. Method references are given in Table~\ref{tab:main}.}
\label{tab:percombo_n3}
\centering
\footnotesize
\renewcommand{\arraystretch}{0.95}
\resizebox{\textwidth}{!}{\begin{tabular}{@{}lrrrrrrrrrrrrrrrrrr@{}}
\toprule
 & \multicolumn{3}{c}{Bl+LL+Hz} & \multicolumn{3}{c}{LL+Hz+Ns} & \multicolumn{3}{c}{Bl+Cl+LL} & \multicolumn{3}{c}{Bl+Cl+LR} & \multicolumn{3}{c}{Cl+Ns+LR} & \multicolumn{3}{c}{Hz+Ns+LR} \\
\cmidrule(lr){2-4} \cmidrule(lr){5-7} \cmidrule(lr){8-10} \cmidrule(lr){11-13} \cmidrule(lr){14-16} \cmidrule(l){17-19}
Method & PSNR & SSIM & LPIPS & PSNR & SSIM & LPIPS & PSNR & SSIM & LPIPS & PSNR & SSIM & LPIPS & PSNR & SSIM & LPIPS & PSNR & SSIM & LPIPS \\
\midrule
Seen & \multicolumn{3}{c}{$\checkmark$} & \multicolumn{3}{c}{$\checkmark$} & \multicolumn{3}{c}{} & \multicolumn{3}{c}{} & \multicolumn{3}{c}{} & \multicolumn{3}{c}{} \\
\midrule
Degraded input & 19.03 & 0.660 & 0.584 & 17.17 & 0.308 & 0.954 & 13.82 & 0.394 & 0.683 & 16.50 & 0.604 & 0.673 & 16.18 & 0.460 & 0.857 & 13.85 & 0.427 & 0.868 \\
\midrule
\multicolumn{19}{@{}l}{\emph{Natural-image all-in-one}} \\
PromptIR & 20.25 & 0.689 & 0.419 & 19.53 & 0.632 & 0.544 & 14.72 & 0.507 & 0.549 & 16.05 & 0.600 & 0.670 & 16.45 & 0.576 & 0.689 & 16.76 & 0.580 & 0.712 \\
Restormer & \underline{22.63} & \underline{0.749} & 0.402 & \underline{21.82} & 0.622 & 0.573 & 16.21 & 0.625 & 0.519 & 15.76 & 0.599 & 0.670 & 15.78 & 0.569 & 0.684 & \underline{22.05} & \underline{0.608} & 0.660 \\
NAFNet & 20.95 & 0.708 & 0.423 & 19.14 & 0.646 & 0.521 & 15.60 & 0.571 & 0.555 & 16.27 & 0.603 & 0.661 & 16.41 & 0.574 & 0.697 & 15.25 & 0.568 & 0.717 \\
DGUNet & 20.25 & 0.694 & 0.407 & 19.63 & 0.613 & 0.552 & 14.22 & 0.457 & 0.513 & 15.90 & 0.598 & 0.668 & 16.30 & 0.560 & 0.714 & 17.62 & 0.556 & 0.748 \\
IDR & 20.09 & 0.677 & 0.460 & 19.99 & 0.530 & 0.602 & 15.83 & 0.575 & 0.572 & 16.15 & 0.592 & 0.680 & 16.52 & 0.571 & 0.708 & 20.02 & 0.524 & 0.787 \\
SRResNet-S+AP & 20.63 & 0.679 & 0.464 & 18.67 & 0.590 & 0.532 & 16.14 & 0.590 & 0.537 & 15.91 & 0.600 & 0.664 & 16.47 & 0.574 & 0.682 & 15.65 & 0.553 & 0.739 \\
Uformer-S+AP & 20.69 & 0.712 & 0.408 & 20.12 & 0.617 & 0.525 & 15.62 & 0.591 & 0.511 & 16.12 & 0.600 & 0.665 & 16.45 & 0.572 & 0.686 & 17.33 & 0.574 & 0.717 \\
Restormer-S+AP & 20.42 & 0.691 & 0.415 & 19.52 & \underline{0.650} & \underline{0.488} & 15.80 & 0.600 & 0.538 & 16.26 & \underline{0.606} & 0.661 & 16.46 & 0.576 & 0.682 & 16.88 & 0.586 & 0.681 \\
AirNet & 20.00 & 0.638 & 0.500 & 20.10 & 0.521 & 0.630 & \underline{18.50} & \underline{0.649} & 0.564 & \underline{16.46} & 0.605 & 0.673 & \underline{16.61} & 0.573 & 0.690 & 19.77 & 0.544 & 0.787 \\
\midrule
\multicolumn{19}{@{}l}{\emph{Remote-sensing all-in-one}} \\
PhyDAE & 20.64 & 0.709 & \underline{0.372} & 19.21 & \textbf{0.664} & \textbf{0.465} & 14.83 & 0.511 & \underline{0.493} & 16.17 & 0.603 & 0.663 & 16.36 & \underline{0.579} & 0.698 & 15.14 & 0.575 & 0.699 \\
Ada4DIR & 21.01 & 0.709 & 0.421 & 18.98 & 0.615 & 0.562 & 16.04 & 0.622 & 0.540 & 15.88 & 0.603 & 0.666 & 16.49 & 0.578 & 0.674 & 16.23 & 0.546 & 0.751 \\
CoRE-UIR & 20.57 & 0.717 & 0.378 & 19.51 & 0.628 & 0.498 & 14.09 & 0.428 & 0.531 & 15.97 & 0.579 & 0.683 & 16.42 & 0.572 & 0.694 & 15.81 & 0.561 & 0.721 \\
\midrule
\multicolumn{19}{@{}l}{\emph{Natural-image systems (zero-shot)}} \\
AutoDIR & 15.55 & 0.435 & 0.665 & 15.18 & 0.512 & 0.587 & 12.05 & 0.425 & 0.611 & 12.23 & 0.454 & 0.811 & 12.92 & 0.402 & 0.885 & 10.83 & 0.357 & 0.961 \\
AgenticIR & 15.46 & 0.537 & 0.554 & 14.62 & 0.457 & 0.775 & 16.34 & 0.591 & 0.514 & 14.19 & 0.357 & \textbf{0.566} & 14.88 & 0.369 & \underline{0.623} & 14.68 & 0.301 & 0.679 \\
4KAgent & 12.59 & 0.349 & 0.581 & 14.22 & 0.489 & 0.594 & 13.65 & 0.430 & 0.512 & 13.86 & 0.347 & 0.601 & 15.96 & 0.453 & \textbf{0.545} & 14.33 & 0.382 & \textbf{0.605} \\
JarvisIR & 11.80 & 0.512 & 0.641 & 11.47 & 0.469 & 0.677 & 13.14 & 0.433 & 0.657 & 13.42 & 0.453 & 0.680 & 13.55 & 0.423 & 0.704 & 10.08 & 0.442 & 0.735 \\
\midrule
\multicolumn{19}{@{}l}{\emph{EORestore-Agent}} \\
Ours & \textbf{23.58} & \textbf{0.755} & \textbf{0.359} & \textbf{21.95} & 0.641 & 0.525 & \textbf{19.03} & \textbf{0.657} & \textbf{0.480} & \textbf{23.25} & \textbf{0.628} & \underline{0.592} & \textbf{22.33} & \textbf{0.601} & 0.636 & \textbf{22.27} & \textbf{0.615} & \underline{0.648} \\
\bottomrule\end{tabular}}
\end{table*}

\begin{table*}[p]
\caption{Per-combination results, $N{=}4$. Column labels: Cl cloud, Hz haze, Ns noise, Bl blur, LL low light, LR low resolution. A check in the Seen row marks combinations composed entirely of degradation types covered by AIO training. Method references are given in Table~\ref{tab:main}.}
\label{tab:percombo_n4}
\centering
\footnotesize
\renewcommand{\arraystretch}{0.95}
\resizebox{\textwidth}{!}{\begin{tabular}{@{}lrrrrrrrrrrrrrrrrrr@{}}
\toprule
 & \multicolumn{3}{c}{Bl+LL+Hz+Ns} & \multicolumn{3}{c}{Bl+Cl+LL+LR} & \multicolumn{3}{c}{Bl+Cl+Hz+LR} & \multicolumn{3}{c}{Bl+LL+Ns+LR} & \multicolumn{3}{c}{Cl+LL+Hz+Ns} & \multicolumn{3}{c}{Cl+Hz+Ns+LR} \\
\cmidrule(lr){2-4} \cmidrule(lr){5-7} \cmidrule(lr){8-10} \cmidrule(lr){11-13} \cmidrule(lr){14-16} \cmidrule(l){17-19}
Method & PSNR & SSIM & LPIPS & PSNR & SSIM & LPIPS & PSNR & SSIM & LPIPS & PSNR & SSIM & LPIPS & PSNR & SSIM & LPIPS & PSNR & SSIM & LPIPS \\
\midrule
Seen & \multicolumn{3}{c}{$\checkmark$} & \multicolumn{3}{c}{} & \multicolumn{3}{c}{} & \multicolumn{3}{c}{} & \multicolumn{3}{c}{} & \multicolumn{3}{c}{} \\
\midrule
Degraded input & 17.33 & 0.254 & 1.033 & 14.21 & 0.389 & 0.741 & 11.94 & 0.514 & 0.758 & 15.51 & 0.313 & 0.874 & 14.52 & 0.295 & 0.937 & 11.95 & 0.385 & 0.904 \\
\midrule
\multicolumn{19}{@{}l}{\emph{Natural-image all-in-one}} \\
PromptIR & 19.29 & 0.576 & 0.663 & 16.39 & 0.588 & 0.676 & 17.35 & \textbf{0.595} & 0.682 & 15.65 & 0.391 & 0.738 & 15.59 & 0.563 & 0.644 & 13.47 & 0.512 & 0.774 \\
Restormer & \underline{20.44} & 0.549 & 0.670 & 16.22 & 0.596 & 0.674 & 16.72 & 0.588 & 0.683 & \underline{22.79} & \underline{0.564} & 0.695 & \underline{17.07} & \textbf{0.600} & \textbf{0.542} & \underline{15.55} & \textbf{0.545} & 0.703 \\
NAFNet & 19.07 & \underline{0.582} & 0.633 & 16.41 & 0.568 & 0.683 & 17.46 & 0.588 & 0.682 & 16.17 & 0.428 & 0.718 & 15.48 & 0.573 & 0.628 & 12.85 & 0.503 & 0.774 \\
DGUNet & 19.44 & 0.555 & 0.649 & 16.13 & 0.586 & 0.678 & 16.58 & 0.582 & 0.692 & 20.74 & 0.539 & 0.730 & 15.59 & 0.544 & 0.633 & 13.54 & 0.489 & 0.803 \\
IDR & 19.93 & 0.462 & 0.703 & 16.35 & 0.568 & 0.688 & \underline{17.65} & \textbf{0.595} & 0.684 & 19.93 & 0.494 & 0.774 & 16.29 & 0.470 & 0.683 & 14.86 & 0.452 & 0.853 \\
SRResNet-S+AP & 18.91 & 0.524 & 0.627 & 16.36 & \textbf{0.598} & \underline{0.669} & 17.22 & 0.591 & 0.675 & 16.29 & 0.429 & 0.740 & 15.32 & 0.521 & 0.623 & 12.82 & 0.481 & 0.796 \\
Uformer-S+AP & 20.12 & 0.551 & 0.635 & 16.25 & 0.593 & 0.674 & 16.50 & 0.584 & 0.686 & 17.08 & 0.489 & 0.708 & 16.04 & 0.551 & 0.610 & 13.45 & 0.503 & 0.780 \\
Restormer-S+AP & 19.59 & 0.581 & 0.617 & 16.60 & \textbf{0.598} & 0.670 & 17.13 & \underline{0.593} & 0.677 & 16.44 & 0.461 & 0.709 & 15.76 & 0.573 & 0.610 & 13.67 & \underline{0.516} & 0.754 \\
AirNet & 19.85 & 0.445 & 0.736 & \underline{16.98} & 0.564 & 0.691 & 17.38 & 0.590 & 0.688 & 18.48 & 0.475 & 0.763 & 16.62 & 0.475 & 0.686 & 14.96 & 0.476 & 0.839 \\
\midrule
\multicolumn{19}{@{}l}{\emph{Remote-sensing all-in-one}} \\
PhyDAE & 19.10 & \textbf{0.593} & \textbf{0.592} & 16.17 & 0.592 & 0.674 & 17.06 & 0.589 & 0.687 & 15.85 & 0.419 & 0.724 & 15.47 & \underline{0.580} & \underline{0.591} & 12.81 & 0.508 & 0.761 \\
Ada4DIR & 19.06 & 0.551 & 0.679 & 16.34 & \underline{0.597} & 0.671 & 16.25 & 0.580 & 0.692 & 16.07 & 0.419 & 0.753 & 15.44 & 0.538 & 0.654 & 13.06 & 0.485 & 0.807 \\
CoRE-UIR & 19.45 & 0.578 & \underline{0.593} & 14.93 & 0.441 & 0.720 & 16.97 & 0.590 & 0.675 & 15.66 & 0.403 & 0.732 & 15.79 & 0.559 & 0.595 & 13.48 & 0.483 & 0.792 \\
\midrule
\multicolumn{19}{@{}l}{\emph{Natural-image systems (zero-shot)}} \\
AutoDIR & 14.53 & 0.426 & 0.681 & 9.59 & 0.372 & 0.889 & 10.80 & 0.449 & 0.823 & 14.30 & 0.363 & 0.898 & 13.68 & 0.429 & 0.655 & 10.40 & 0.355 & 0.965 \\
AgenticIR & 15.28 & 0.463 & 0.789 & 15.08 & 0.302 & \textbf{0.637} & 13.17 & 0.307 & \textbf{0.636} & 17.29 & 0.239 & 0.735 & 13.87 & 0.462 & 0.742 & 12.95 & 0.325 & \underline{0.700} \\
4KAgent & 12.85 & 0.465 & 0.712 & 14.07 & 0.263 & 0.688 & 13.36 & 0.316 & \underline{0.650} & 14.91 & 0.326 & \textbf{0.598} & 13.39 & 0.476 & 0.681 & 12.22 & 0.368 & \textbf{0.638} \\
JarvisIR & 10.32 & 0.447 & 0.744 & 13.17 & 0.410 & 0.699 & 10.37 & 0.430 & 0.722 & 14.12 & 0.438 & 0.707 & 10.62 & 0.438 & 0.727 & 10.12 & 0.424 & 0.772 \\
\midrule
\multicolumn{19}{@{}l}{\emph{EORestore-Agent}} \\
Ours & \textbf{21.07} & 0.572 & 0.626 & \textbf{18.93} & 0.592 & \textbf{0.637} & \textbf{18.76} & 0.515 & 0.665 & \textbf{26.08} & \textbf{0.630} & \underline{0.655} & \textbf{19.57} & 0.528 & 0.597 & \textbf{18.95} & 0.461 & 0.716 \\
\bottomrule\end{tabular}}
\end{table*}

\begin{table*}[p]
\caption{Per-combination results, $N{=}5$. Column labels: Cl cloud, Hz haze, Ns noise, Bl blur, LL low light, LR low resolution. A check in the Seen row marks combinations composed entirely of degradation types covered by AIO training. Method references are given in Table~\ref{tab:main}.}
\label{tab:percombo_n5}
\centering
\footnotesize
\renewcommand{\arraystretch}{0.95}
\resizebox{\textwidth}{!}{\begin{tabular}{@{}lrrrrrrrrrrrrrrrrrr@{}}
\toprule
 & \multicolumn{3}{c}{Bl+Cl+LL+Hz+Ns} & \multicolumn{3}{c}{Bl+Cl+LL+Hz+LR} & \multicolumn{3}{c}{Bl+Cl+LL+Ns+LR} & \multicolumn{3}{c}{Bl+Cl+Hz+Ns+LR} & \multicolumn{3}{c}{Bl+LL+Hz+Ns+LR} & \multicolumn{3}{c}{Cl+LL+Hz+Ns+LR} \\
\cmidrule(lr){2-4} \cmidrule(lr){5-7} \cmidrule(lr){8-10} \cmidrule(lr){11-13} \cmidrule(lr){14-16} \cmidrule(l){17-19}
Method & PSNR & SSIM & LPIPS & PSNR & SSIM & LPIPS & PSNR & SSIM & LPIPS & PSNR & SSIM & LPIPS & PSNR & SSIM & LPIPS & PSNR & SSIM & LPIPS \\
\midrule
Seen & \multicolumn{3}{c}{} & \multicolumn{3}{c}{} & \multicolumn{3}{c}{} & \multicolumn{3}{c}{} & \multicolumn{3}{c}{} & \multicolumn{3}{c}{} \\
\midrule
Degraded input & 14.74 & 0.226 & 1.030 & 14.91 & 0.529 & 0.791 & 14.09 & 0.282 & 0.897 & 11.79 & 0.355 & 0.917 & 17.48 & 0.401 & 0.922 & 14.92 & 0.381 & 0.931 \\
\midrule
\multicolumn{19}{@{}l}{\emph{Natural-image all-in-one}} \\
PromptIR & 15.93 & 0.518 & 0.725 & 16.07 & 0.504 & 0.736 & 14.04 & 0.342 & 0.806 & 13.45 & 0.491 & 0.799 & 19.45 & 0.533 & 0.781 & 16.00 & 0.501 & 0.798 \\
Restormer & \underline{17.27} & \textbf{0.543} & \textbf{0.632} & \underline{17.02} & \textbf{0.554} & 0.719 & \underline{16.24} & \underline{0.509} & 0.754 & \underline{15.20} & \textbf{0.520} & 0.739 & \underline{20.10} & 0.530 & 0.787 & \underline{16.90} & \textbf{0.525} & 0.749 \\
NAFNet & 15.82 & 0.518 & 0.705 & 16.24 & 0.508 & 0.737 & 14.15 & 0.355 & 0.800 & 12.68 & 0.478 & 0.801 & 18.67 & 0.550 & 0.783 & 15.71 & 0.511 & 0.796 \\
DGUNet & 15.93 & 0.492 & 0.708 & 16.05 & 0.514 & 0.742 & 15.19 & 0.423 & 0.805 & 13.55 & 0.471 & 0.820 & 19.68 & 0.525 & 0.791 & 16.06 & 0.487 & 0.815 \\
IDR & 16.41 & 0.406 & 0.766 & 15.84 & 0.473 & 0.738 & 14.56 & 0.362 & 0.812 & 14.74 & 0.428 & 0.860 & 19.84 & 0.478 & 0.810 & 16.33 & 0.449 & 0.837 \\
SRResNet-S+AP & 15.71 & 0.459 & 0.683 & 16.05 & 0.503 & 0.740 & 14.41 & 0.380 & 0.799 & 12.80 & 0.457 & 0.821 & 18.71 & 0.512 & 0.802 & 15.43 & 0.481 & 0.812 \\
Uformer-S+AP & 16.19 & 0.491 & 0.693 & 15.97 & \underline{0.525} & 0.736 & 14.42 & 0.396 & 0.795 & 13.45 & 0.481 & 0.809 & 19.59 & 0.517 & 0.787 & 16.12 & 0.495 & 0.806 \\
Restormer-S+AP & 16.01 & 0.515 & 0.704 & 15.96 & 0.500 & 0.731 & 14.32 & 0.379 & 0.787 & 13.71 & \underline{0.495} & 0.783 & 19.90 & \underline{0.554} & 0.767 & 16.26 & 0.510 & 0.785 \\
AirNet & 16.61 & 0.409 & 0.766 & 16.04 & 0.477 & 0.750 & 14.89 & 0.389 & 0.807 & 14.84 & 0.452 & 0.852 & 19.78 & 0.463 & 0.822 & 16.36 & 0.438 & 0.844 \\
\midrule
\multicolumn{19}{@{}l}{\emph{Remote-sensing all-in-one}} \\
PhyDAE & 15.79 & \underline{0.525} & 0.678 & 16.11 & 0.522 & 0.737 & 14.23 & 0.364 & 0.800 & 12.90 & 0.481 & 0.791 & 18.61 & 0.553 & 0.774 & 15.58 & \underline{0.513} & 0.789 \\
Ada4DIR & 15.69 & 0.492 & 0.730 & 15.80 & 0.506 & 0.736 & 14.25 & 0.362 & 0.803 & 12.99 & 0.465 & 0.825 & 18.87 & 0.511 & 0.803 & 15.64 & 0.483 & 0.809 \\
CoRE-UIR & 15.96 & 0.518 & 0.656 & 16.46 & 0.523 & 0.732 & 14.14 & 0.352 & 0.806 & 13.37 & 0.461 & 0.811 & 19.04 & 0.538 & 0.796 & 15.83 & 0.497 & 0.811 \\
\midrule
\multicolumn{19}{@{}l}{\emph{Natural-image systems (zero-shot)}} \\
AutoDIR & 14.09 & 0.420 & 0.707 & 12.87 & 0.453 & 0.855 & 12.42 & 0.321 & 0.910 & 10.14 & 0.340 & 0.976 & 11.87 & 0.331 & 0.989 & 11.65 & 0.346 & 0.986 \\
AgenticIR & 14.63 & 0.458 & 0.791 & 13.60 & 0.304 & \textbf{0.672} & 13.72 & 0.245 & \underline{0.720} & 12.39 & 0.343 & \underline{0.688} & 14.91 & 0.312 & \underline{0.737} & 14.42 & 0.349 & \underline{0.730} \\
4KAgent & 13.60 & 0.436 & 0.712 & 14.30 & 0.331 & \underline{0.691} & 13.31 & 0.282 & \textbf{0.635} & 11.65 & 0.371 & \textbf{0.646} & 14.32 & 0.415 & \textbf{0.700} & 14.91 & 0.433 & \textbf{0.644} \\
JarvisIR & 10.23 & 0.411 & 0.763 & 9.99 & 0.419 & 0.762 & 12.61 & 0.391 & 0.780 & 10.38 & 0.417 & 0.795 & 10.10 & 0.427 & 0.829 & 10.13 & 0.413 & 0.810 \\
\midrule
\multicolumn{19}{@{}l}{\emph{EORestore-Agent}} \\
Ours & \textbf{19.79} & 0.499 & \underline{0.652} & \textbf{19.20} & 0.513 & 0.702 & \textbf{18.40} & \textbf{0.520} & 0.722 & \textbf{18.70} & 0.464 & 0.727 & \textbf{20.89} & \textbf{0.556} & 0.773 & \textbf{19.66} & 0.500 & \underline{0.730} \\
\bottomrule\end{tabular}}
\end{table*}

\begin{table*}[p]
\caption{Per-combination results, $N{=}6$. The four columns apply all six types in different orders: Order 1: Bl$\rightarrow$Cl$\rightarrow$LL$\rightarrow$Hz$\rightarrow$Ns$\rightarrow$LR, Order 2: Ns$\rightarrow$Bl$\rightarrow$Cl$\rightarrow$LR$\rightarrow$LL$\rightarrow$Hz, Order 3: Ns$\rightarrow$LL$\rightarrow$Cl$\rightarrow$Bl$\rightarrow$LR$\rightarrow$Hz, Order 4: LR$\rightarrow$Hz$\rightarrow$Bl$\rightarrow$LL$\rightarrow$Cl$\rightarrow$Ns. Column labels: Cl cloud, Hz haze, Ns noise, Bl blur, LL low light, LR low resolution. Method references are given in Table~\ref{tab:main}.}
\label{tab:percombo_n6}
\centering
\footnotesize
\renewcommand{\arraystretch}{0.95}
\resizebox{0.68\textwidth}{!}{\begin{tabular}{@{}lrrrrrrrrrrrr@{}}
\toprule
 & \multicolumn{3}{c}{Order 1} & \multicolumn{3}{c}{Order 2} & \multicolumn{3}{c}{Order 3} & \multicolumn{3}{c}{Order 4} \\
\cmidrule(lr){2-4} \cmidrule(lr){5-7} \cmidrule(lr){8-10} \cmidrule(l){11-13}
Method & PSNR & SSIM & LPIPS & PSNR & SSIM & LPIPS & PSNR & SSIM & LPIPS & PSNR & SSIM & LPIPS \\
\midrule
Seen & \multicolumn{3}{c}{} & \multicolumn{3}{c}{} & \multicolumn{3}{c}{} & \multicolumn{3}{c}{} \\
\midrule
Degraded input & 14.37 & 0.362 & 0.938 & 14.97 & 0.519 & 0.818 & 14.14 & 0.509 & 0.840 & 14.32 & 0.257 & 0.912 \\
\midrule
\multicolumn{13}{@{}l}{\emph{Natural-image all-in-one}} \\
PromptIR & 15.75 & 0.484 & 0.821 & 16.08 & 0.494 & 0.765 & 15.45 & 0.474 & 0.783 & 15.18 & 0.461 & 0.868 \\
Restormer & \underline{16.66} & \textbf{0.508} & 0.774 & \underline{16.89} & \textbf{0.542} & 0.736 & \underline{16.11} & \textbf{0.529} & 0.754 & 15.76 & \underline{0.472} & 0.838 \\
NAFNet & 15.33 & 0.492 & 0.820 & 16.35 & 0.505 & 0.762 & 15.74 & 0.488 & 0.782 & 14.95 & 0.453 & 0.858 \\
DGUNet & 15.72 & 0.470 & 0.830 & 16.14 & 0.509 & 0.765 & 15.40 & 0.489 & 0.780 & 14.81 & 0.443 & 0.858 \\
IDR & 16.12 & 0.424 & 0.854 & 16.10 & 0.474 & 0.765 & 15.63 & 0.467 & 0.778 & 15.27 & 0.407 & 0.857 \\
SRResNet-S+AP & 15.06 & 0.460 & 0.840 & 15.76 & 0.469 & 0.772 & 15.41 & 0.452 & 0.785 & 14.70 & 0.446 & 0.861 \\
Uformer-S+AP & 15.76 & 0.477 & 0.829 & 16.13 & 0.520 & 0.760 & 15.54 & \underline{0.505} & 0.776 & 14.99 & 0.450 & 0.852 \\
Restormer-S+AP & 15.95 & \underline{0.493} & 0.810 & 16.09 & 0.499 & 0.755 & 15.53 & 0.482 & 0.769 & 15.28 & 0.464 & 0.861 \\
AirNet & 16.24 & 0.420 & 0.854 & 16.26 & 0.480 & 0.772 & 15.77 & 0.464 & 0.789 & \underline{15.85} & 0.417 & 0.839 \\
\midrule
\multicolumn{13}{@{}l}{\emph{Remote-sensing all-in-one}} \\
PhyDAE & 15.25 & 0.486 & 0.822 & 16.05 & 0.506 & 0.766 & 15.55 & 0.490 & 0.781 & 14.97 & 0.464 & 0.875 \\
Ada4DIR & 15.28 & 0.466 & 0.834 & 15.80 & 0.496 & 0.772 & 15.14 & 0.480 & 0.789 & 14.91 & 0.458 & 0.859 \\
CoRE-UIR & 15.42 & 0.474 & 0.838 & 16.52 & 0.514 & 0.760 & 15.77 & 0.504 & 0.779 & 14.78 & 0.446 & 0.882 \\
\midrule
\multicolumn{13}{@{}l}{\emph{Natural-image systems (zero-shot)}} \\
AutoDIR & 11.33 & 0.340 & 0.988 & 13.50 & 0.474 & 0.852 & 12.70 & 0.460 & 0.876 & 11.55 & 0.300 & 0.978 \\
AgenticIR & 13.48 & 0.343 & \underline{0.709} & 13.65 & 0.303 & \underline{0.696} & 13.36 & 0.311 & \underline{0.684} & 13.43 & 0.308 & \underline{0.735} \\
4KAgent & 13.74 & 0.424 & \textbf{0.657} & 14.52 & 0.367 & \textbf{0.653} & 13.27 & 0.334 & \textbf{0.674} & 13.24 & 0.362 & \textbf{0.699} \\
JarvisIR & 9.88 & 0.403 & 0.829 & 10.32 & 0.413 & 0.756 & 10.19 & 0.416 & 0.780 & 11.15 & 0.376 & 0.876 \\
\midrule
\multicolumn{13}{@{}l}{\emph{EORestore-Agent}} \\
Ours & \textbf{19.26} & 0.491 & 0.755 & \textbf{19.68} & \underline{0.522} & 0.703 & \textbf{19.02} & \underline{0.505} & 0.716 & \textbf{17.96} & \textbf{0.483} & 0.806 \\
\bottomrule\end{tabular}}
\end{table*}}

\clearpage

\bibliographystyle{IEEEtran}
\bibliography{refs}

\newpage

 
\vspace{11pt}

\begin{IEEEbiography}[{\includegraphics[width=1in,height=1.25in,clip,keepaspectratio]{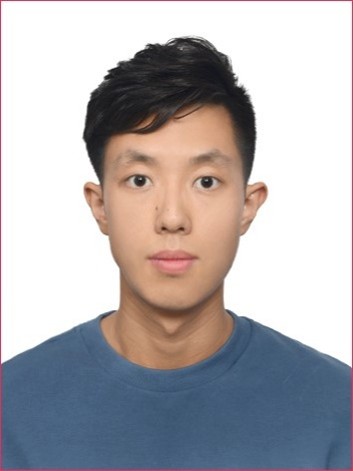}}]{Heli Qi}
received the B.Eng. degree in software engineering from Northeastern University, Shenyang, China, in 2020, and the M.Eng. degree in information science and engineering from Nara Institute of Science and Technology (NAIST), Ikoma, Japan, in 2024. He is currently pursuing the D.Eng. degree in information architecture with the Graduate School of Information, Production and Systems, Waseda University, Kitakyushu, Japan. His research interests include multimodal learning, large language model evaluation, AI agents, and remote sensing.

He is also working as a part-time researcher with the Geoinformatics Team, RIKEN Center for Advanced Intelligence Project (AIP), Tokyo, Japan. His work has been published at major conferences including NeurIPS, ICML, ACL, EMNLP, ECCV, and INTERSPEECH. He was a member of the DisasterM3 team that won first place in the 2025 ESA \mbox{$\Phi$-lab} AI for Earthquake Response Challenge.
\end{IEEEbiography}

\vspace{11pt}
\begin{IEEEbiography}[{\includegraphics[width=1in,height=1.25in,clip,keepaspectratio]{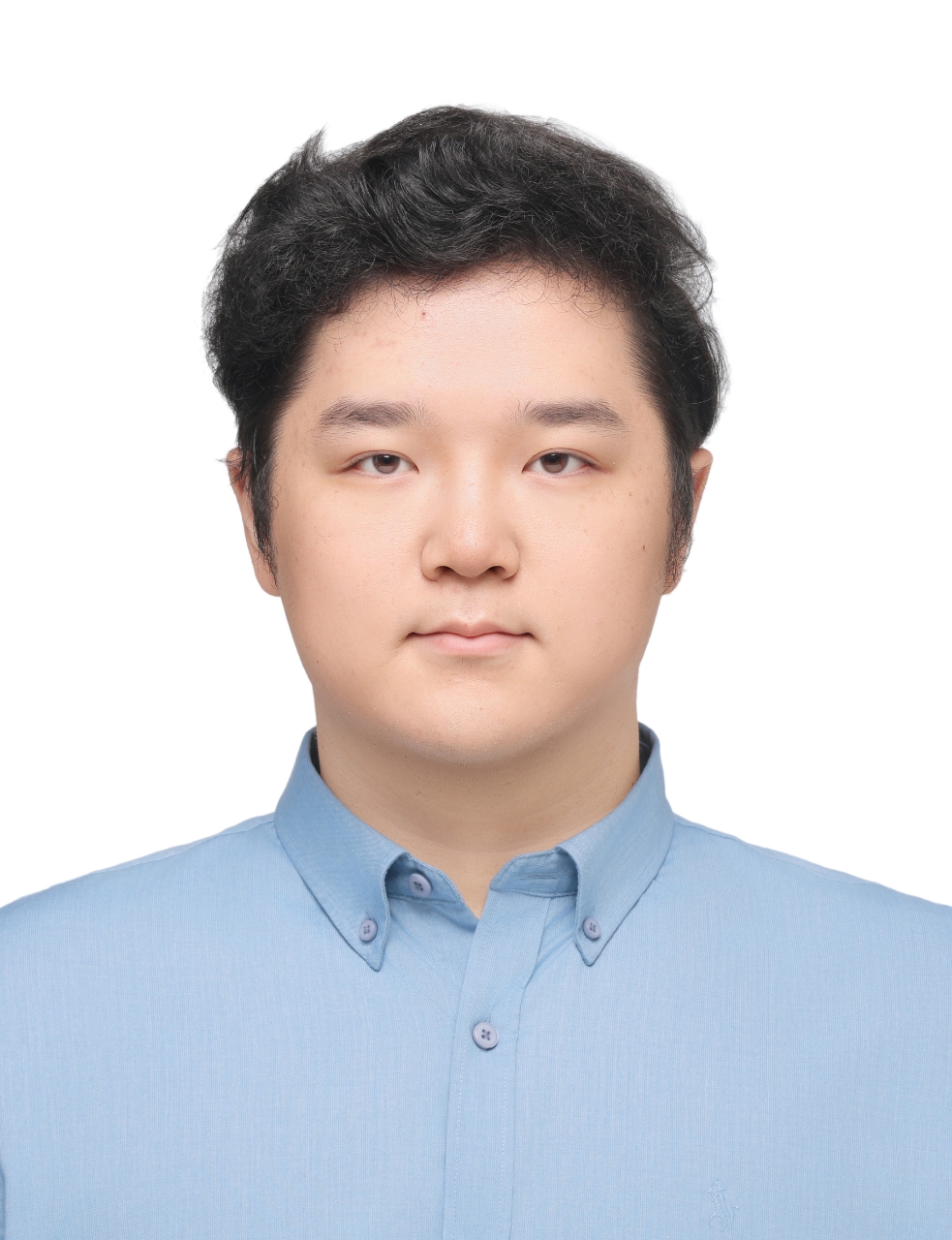}}]{Zeqi Zhou}
received dual B.S. degrees in computer science and statistics and analytics from the University of North Carolina at Chapel Hill, Chapel Hill, NC, USA, in 2024, and the M.Sc. degree in computer science from Brown University, Providence, RI, USA, in 2026. He is currently pursuing the Ph.D. degree with the Department of Complexity Science and Engineering, The University of Tokyo, Tokyo, Japan. His research focuses on agents for remote sensing.

His work has been accepted by the International Conference on Machine Learning (ICML 2026), the Annual Meeting of the Association for Computational Linguistics (ACL 2026), and the Conference on Empirical Methods in Natural Language Processing (EMNLP 2026).
\end{IEEEbiography}

\vspace{11pt}
\begin{IEEEbiography}[{\includegraphics[width=1in,height=1.25in,clip,keepaspectratio]{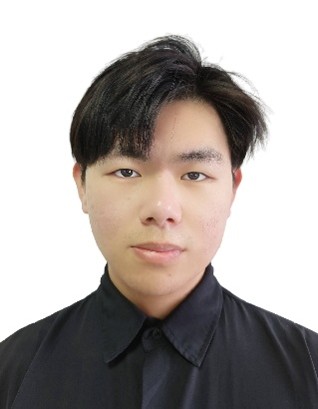}}]{Jingjun Yi}
received the B.S. degree from the School of Remote Sensing and Information Engineering, Wuhan University, Wuhan, China, in 2023, and the M.S. degree from the State Key Laboratory of Information Engineering in Surveying, Mapping and Remote Sensing, Wuhan University, Wuhan, China, in 2025. He is currently pursuing the Ph.D. degree in computer vision and machine learning at the University of Amsterdam, Amsterdam, The Netherlands. His research focuses on computer vision and machine learning, including hyperbolic deep learning, multimodal large language models, and generative models.
\end{IEEEbiography}

\vspace{11pt}
\begin{IEEEbiography}[{\includegraphics[width=1in,height=1.25in,clip,keepaspectratio]{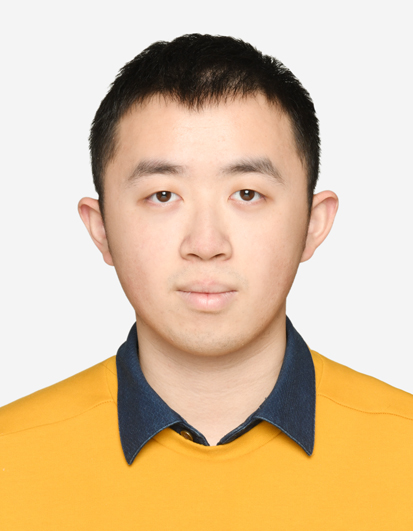}}]{Kunyi Liu}
received the B.E. degree in software engineering from Dalian University of Technology, Dalian, China, in 2021, and the M.E. degree in information, production and systems engineering from the Graduate School of Information, Production and Systems, Waseda University, Kitakyushu, Japan, in 2024, where he is currently pursuing the Ph.D. degree.

He is also working as a part-time researcher with the Geoinformatics Team, RIKEN Center for Advanced Intelligence Project (AIP), Tokyo, Japan. His research interests include image processing, deep learning, agentic artificial intelligence, and their application in Earth observation.
\end{IEEEbiography}

\vspace{11pt}

\begin{IEEEbiography}[{\includegraphics[width=1in,height=1.25in,clip,keepaspectratio]{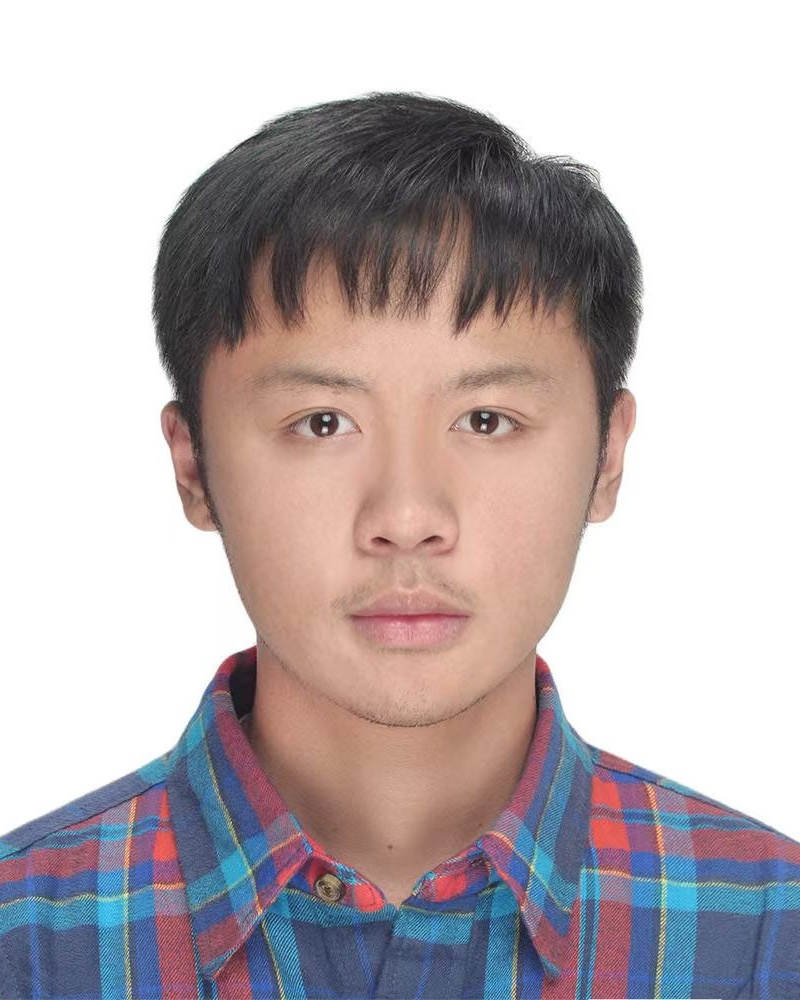}}]{Ziyang Lihe}
(Student Member, IEEE) received the B.Eng. degree in geomatics engineering from Southwest Jiaotong University, Chengdu, China, in 2022, and the M.Eng. degree in resources and environment from Wuhan University, Wuhan, China, in 2024. He is currently working toward the Ph.D. degree in photogrammetry and remote sensing with Wuhan University, Wuhan, China.

His research mainly focuses on remote sensing image restoration, single-image all-in-one restoration, and satellite image time-series reconstruction.
\end{IEEEbiography}

\vspace{11pt}
\begin{IEEEbiography}[{\includegraphics[width=1in,height=1.25in,clip,keepaspectratio]{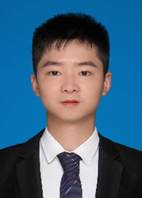}}]{Junjue Wang}
received the B.S. degree from the School of Geography and Information Engineering, China University of Geosciences, Wuhan, China, in 2019 and the doctoral degree in photogrammetry and remote sensing from Wuhan University, Wuhan, in 2024. He is currently an Assistant Professor at the Department of Complexity Science and Engineering, The University of Tokyo. His major research interests are multimodal remote sensing data processing.

He won the first-place prize in 2025 ESA \mbox{$\Phi$-lab} AI for Earthquake Response Challenge, 2022 LandSlide4Sense Contest, the second-place prize in Single-view Semantic 3D Challenge of 2019 IEEE GRSS Data Fusion Contest, the fourth-place in xView2 Challenge.
\end{IEEEbiography}

\vspace{11pt}
\begin{IEEEbiography}[{\includegraphics[width=1in,height=1.25in,clip,keepaspectratio]{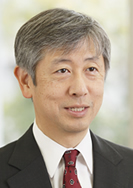}}]{Osamu Yoshie}
(Member, IEEE) is currently a Professor with Waseda University, Shinjuku, Japan, where he is the Director of the Institute for Global Strategies on Industry-Academia Fusion (IGSIAF). He has held research positions with FX Palo Alto Laboratory, USA, and Fraunhofer (formerly, GMD), Germany. His research interests include computer vision, augmented reality, artificial intelligence, industrial IoT, and community analysis and synthesis. He is a fellow of the Institute of Electrical Engineers of Japan (IEEJ).
\end{IEEEbiography}

\vspace{11pt}
\begin{IEEEbiography}[{\includegraphics[width=1in,height=1.25in,clip,keepaspectratio]{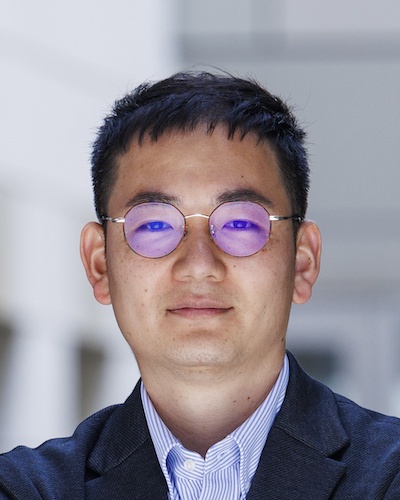}}]{Naoto Yokoya}
(Member, IEEE) received the M.Eng. and Ph.D. degrees in aerospace engineering from The University of Tokyo, Tokyo, Japan, in 2010 and 2013, respectively. He is currently a Professor with the Graduate School of Frontier Sciences, The University of Tokyo, and the Team Director of the Geoinformatics Team at the RIKEN Center for Advanced Intelligence Project, Tokyo, Japan. He was an Alexander von Humboldt Fellow at the German Aerospace Center (DLR) and the Technical University of Munich (TUM) from 2015 to 2017.

His research interests include remote sensing, computer vision, image processing, data fusion, and machine learning, with applications to disaster management and environmental assessment. He serves as an Associate Editor for the \textsc{IEEE Transactions on Pattern Analysis and Machine Intelligence} and the \textit{ISPRS Journal of Photogrammetry and Remote Sensing}. He has been recognized as a Clarivate Highly Cited Researcher every year since 2022.
\end{IEEEbiography}


\vfill

\end{document}